%% file: arxiv.tex
\documentclass{article}

\usepackage[preprint]{neurips_2026}

\usepackage[utf8]{inputenc} 
\usepackage[T1]{fontenc}    
\usepackage{hyperref}       
\usepackage{url}            
\usepackage{booktabs}       
\usepackage{amsfonts}       
\usepackage{nicefrac}       
\usepackage{microtype}      
\usepackage{xcolor}         

\usepackage{multirow}

\input{math_commands}

\usepackage{graphicx}
\usepackage{subcaption}
\usepackage{wrapfig}
\usepackage{pifont} 
\usepackage{colortbl}
\usepackage{soul}
\usepackage{newfloat}
\usepackage{listings}
\usepackage{algorithm}
\usepackage{algorithmic}
\usepackage{enumitem}
\usepackage{lineno}
\usepackage{graphicx}
\usepackage{textcomp}
\usepackage{multirow}
\usepackage{multicol}
\usepackage{color}
\usepackage{bm}
\usepackage{arydshln}
\usepackage{makecell}
\usepackage[textsize=tiny]{todonotes}
\usepackage{cleveref}
\usepackage[most]{tcolorbox}
\definecolor{lightgraybox}{gray}{0.95}

\hypersetup{
    colorlinks=true,
    citecolor=blue,
    linkcolor=blue,
    urlcolor=blue
}

\newcommand{\cmark}{\textcolor{green!60!black}{\ding{51}}} 
\newcommand{\xmark}{\textcolor{red!70!black}{\ding{55}}}   
\newcommand{\nmark}{\textcolor{orange!85!black}{\ding{108}}} 

\ifodd 1
\newcommand{\czh}[1]{{\color{cyan}(czh: #1)}}
\newcommand{\cssc}[1]{{\color{orange}[Chengshuai: {#1}]}}

\else
\newcommand{\czh}[1]{}
\newcommand{\cssc}[1]{}
\fi

\newcommand{\algcomment}[1]{\hfill $\triangleright$~#1}
\newcommand{\up}[1]{\textsubscript{\textcolor{red}{\scriptsize$\uparrow$#1}}}
\newcommand{\down}[1]{\textsubscript{\textcolor{blue}{\scriptsize$\downarrow$#1}}}
\title{Is One Score Enough? Rethinking the Evaluation\\ of Sequentially Evolving LLM Memory}

\author{%
    Songwei Dong\thanks{indicates equal contributions, random order.} \\
    University of Virginia\\
  \texttt{hxt5ap@virginia.edu} \\
  \And
  Zihan Chen$^*$ \\
  University of Virginia\\
  \texttt{brf3rx@virginia.edu} \\
  \And
  Chengshuai Shi \\
  Princeton University\\
  \texttt{cs1083@princeton.edu} \\
  \And
  Peng Wang \\
  University of Virginia\\
  \texttt{pw7nc@virginia.edu} \\
  \And
  Jundong Li \\
  University of Virginia\\
  \texttt{jundong@virginia.edu} \\
  \And
  Cong Shen \\
  University of Virginia \\
  \texttt{cong@virginia.edu} \\
}

\begin{document}

\maketitle

\begin{abstract}
Memory plays a central role in enabling large language models (LLMs) to operate over sequential tasks by accumulating and reusing experience over time. However, existing evaluations of LLM memory mostly rely on aggregate metrics such as final hold-out accuracy or cumulative online performance. We argue that these metrics can be misleading: they collapse distinct memory behaviors into a single number and obscure critical failure modes such as forgetting and negative transfer. In this paper, we introduce \textsc{SeqMem-Eval}, a diagnostic evaluation framework for sequentially evolving LLM memory. Drawing inspiration from continual learning, it targets a distinct test-time setting in which memory is external, prompt-mediated, and updated without changing model parameters. Rather than only measuring whether the final memory state improves performance, \textsc{SeqMem-Eval} examines how memory states evolve, generalize, consolidate experience, and retain useful information during sequential inference. Specifically, it measures online utility, hold-out generalization, backward transfer, and forgetting, providing a finer-grained view of whether memory updates help current tasks, generalize to unseen tasks, improve past predictions, or degrade previously acquired knowledge.
Through extensive experiments across diverse tasks and memory methods, we uncover several previously overlooked phenomena. In particular, we show that higher final or cumulative accuracy does not necessarily imply better memory quality: many methods exhibit strong performance gains while suffering from significant forgetting or negative transfers. Moreover, different memory designs exhibit distinct trade-offs between adaptability and stability, which are invisible under standard evaluation metrics. Our findings show that aggregate metrics systematically miss several recurring failure modes, suggesting that a multi-dimensional perspective is essential for understanding LLM memory. Code and evaluation framework are available at:
\url{https://github.com/ShenGroup/SeqMem-Eval}
\end{abstract}

\section{Introduction}

Large language models (LLMs) are increasingly equipped with external memory to evolve over sequential tasks~\citep{xiang2026systematic,fang2025memp,wei2025evo}, where models are expected to accumulate experience and adapt their behavior over time. 
Through continual interaction with tasks or environments, LLMs generate rich trajectories that contain not only successful solutions but also failed attempts, feedback signals, and intermediate reasoning traces~\citep{zhao2024expelllmagentsexperiential}. 
Rather than serving as passive records of past interactions, these trajectories can provide valuable experience for refining future decisions, improving task strategies, and enabling test-time adaptation~\citep{wei2025evo,suzgun2025dynamiccheatsheettesttimelearning,zhou2025memento}.
Such sequentially evolving systems are central to many emerging applications, including reasoning assistants~\citep{ho2025arcmemo}, tool-use agents~\citep{wang2025reinforcement}, and interactive decision-making systems~\citep{zheng2025skillweaver,agrawal2025gepa}, where performance depends not only on the current input but also on previously encountered tasks. Wei et al.~\cite{wei2025evo} formalize this setting by viewing memory as an evolving state that is \textit{retrieved}, \textit{synthesized}, and \textit{updated} throughout a task sequence. This perspective marks an important step beyond static conversational recall, highlighting the role of memory in test-time adaptation and experience reuse.

Despite this progress, the evaluation of sequentially evolving LLM memory remains incomplete. Existing studies typically assess memory methods using aggregate performance metrics, such as final hold-out accuracy after memory construction~\citep{zhao2024expelllmagentsexperiential} or cumulative online accuracy along the sequence~\citep{wei2025evo,suzgun2025dynamiccheatsheettesttimelearning}. While useful, these metrics collapse the behavior of an evolving memory system into a single number. As a result, they do not reveal whether memory updates genuinely improve future behavior, whether later experiences help consolidate earlier ones, or whether the system forgets knowledge that was previously useful. In practice, similar final or average accuracy may mask fundamentally different learning dynamics: one method may steadily accumulate reusable knowledge, while another may exhibit oscillatory behavior or transient improvements followed by degradation. Thus, aggregate metrics can create an illusion of comparable memory quality, even when the underlying memory dynamics are substantially different.

\begin{figure}[t]
    \centering
    \includegraphics[width = \textwidth]{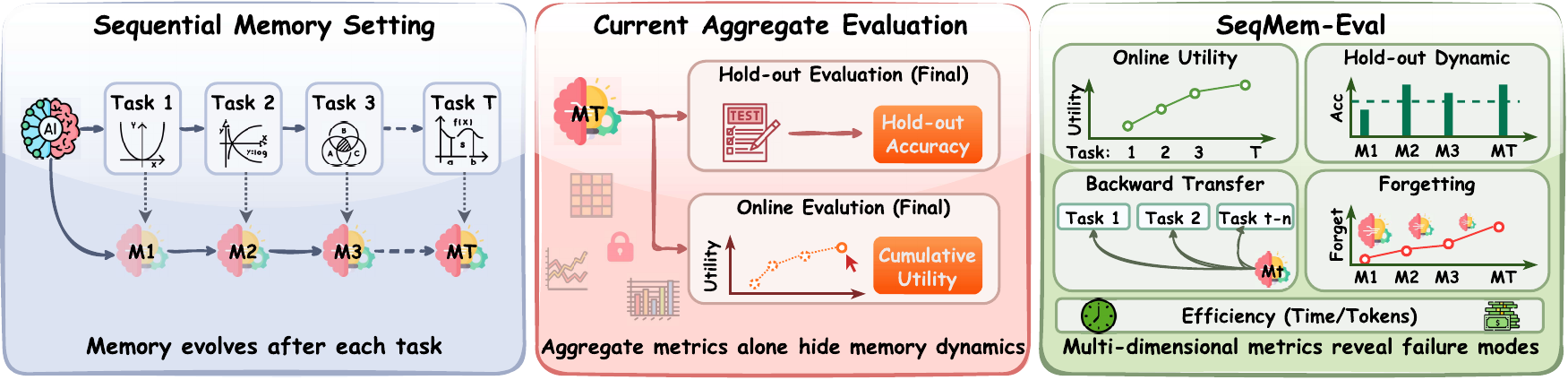}
    \caption{\textbf{\textsc{SeqMem-Eval}: Beyond aggregate evaluation of LLM memory.}
Left: In sequential settings, an LLM processes a stream of tasks while maintaining an evolving memory state.
Middle: Existing evaluations reduce memory performance to aggregate metrics, which collapses complex memory dynamics and hides important behaviors.
Right: \textsc{SeqMem-Eval} decomposes memory quality into multiple dimensions, including online utility, hold-out generalization, backward transfer, forgetting, and efficiency, enabling fine-grained analysis of how memory evolves. 
}
\vspace{-0.2in}
    \label{fig:teaser}
\end{figure}

In this paper, we propose \textsc{SeqMem-Eval}, a diagnostic evaluation framework for sequentially evolving LLM memory. While motivated by continual learning~\citep{wu2022pretrained,lopez2017gradient}, sequentially evolving LLM memory is a distinct test-time setting: the LLM remains fixed, and adaptation occurs through updates to an external textual memory that influences predictions via retrieval and context construction.  Therefore, evaluation should focus not only on final task performance, but also on how these memory updates affect predictions throughout the sequence.  \textsc{SeqMem-Eval} captures this behavior through five complementary dimensions:  \emph{online utility}, \emph{hold-out generalization}, \emph{backward transfer}, \emph{forgetting}, and \emph{efficiency}. Together, these diagnostics reveal whether memory updates are useful, transferable, stable, and cost-effective, rather than merely improving aggregate accuracy.

We conduct a systematic empirical study across diverse tasks, models, and representative memory methods under the \textsc{SeqMem-Eval} protocol. Our results show that standard aggregate metrics can be misleading: methods with strong final or online accuracy may still exhibit substantial forgetting, limited backward transfer, or weak generalization from accumulated experience. These findings suggest that current evaluation practices can overestimate the effectiveness of LLM memory and obscure important failure modes. Our contributions are summarized as follows:
\begin{itemize}[leftmargin=0.35cm]
    \item \textbf{Diagnostic evaluation framework.} We introduce \textsc{SeqMem-Eval}, a continual-learning-inspired framework for evaluating sequentially evolving LLM memory beyond aggregate accuracy.
    \item \textbf{Comprehensive empirical study.} We provide a systematic comparison of representative memory methods across diverse tasks and models under a unified sequential evaluation protocol.
    \item \textbf{Actionable findings for memory design.} We identify key failure modes, including forgetting, limited backward transfer, and weak generalization, offering design implications for more reliable memory-augmented LLMs.
\end{itemize}

\section{Related Work}

\paragraph{Sequentially evolving LLM memory and evaluation.}
Memory has become a central mechanism for enabling LLM agents to move beyond isolated inputs and adapt over sequential interactions~\citep{xiang2026systematic,fang2025comprehensive,madaan2023self,chhikara2025mem0}. Recent memory-augmented agents extract reusable information from prior trajectories, feedback, reflections, or workflows, and use such information to improve future reasoning and decision-making~\citep{wang2025far,chen2025swe,fang2025memp,zhong2024memorybank,xu2025mem}. Representative approaches include reflection- or experience-based methods such as ExpeL~\citep{zhao2024expelllmagentsexperiential}, dynamic memory construction methods such as Dynamic Cheatsheet~\citep{suzgun2025dynamiccheatsheettesttimelearning}, workflow-level memory methods such as Agent Workflow Memory (AWM)~\citep{wang2024agentworkflowmemory}, and retrieval- or structure-based memory systems such as G-Memory~\citep{zhang2025gmemorytracinghierarchicalmemory} and Memento~\citep{zhou2025memento}. These methods differ in how they store, retrieve, and update memory, but share the goal of improving future behavior through test-time experience reuse~\citep{tang2025agent,feng2025get,ho2025arcmemo}. Recent benchmark efforts further formalize this setting: for example, Evo-Memory~\citep{wei2025evo} converts static datasets into sequential task streams and evaluates agents whose memory is searched, synthesized, and evolved after each interaction. Broader works of self-evolving agents also frame memory and trajectory reuse as part of environment-centric self-evolution~\cite{xiang2026systematic,gao2025survey}. However, existing evaluations still largely rely on aggregate metrics such as final hold-out accuracy, cumulative online accuracy, or average success rate~\citep{ouyang2025reasoningbank,wu2025evolver}. Such metrics are useful for comparing end performance, but they provide limited insight into memory dynamics, like whether a method retains useful information. Our work complements prior memory methods and benchmarks by focusing on diagnostic evaluation rather than proposing another memory architecture.

\paragraph{Continual learning evaluation and diagnostic metrics.}
Our evaluation perspective is inspired by continual learning, where models learn from a sequence of tasks while attempting to acquire new knowledge without forgetting old knowledge~\citep{biesialska2020continual,kirkpatrick2017overcoming,wang2024comprehensive}. Classical continual learning evaluation uses metrics such as backward transfer, forward transfer, and stability--plasticity trade-offs to characterize learning dynamics beyond final accuracy~\citep{chaudhry2018efficient,chaudhry2019tiny,lopez2017gradient,wu2022pretrained,qi2023fine,wang2023trace}. Recent works on continual learning for LLMs further highlight the importance of updating large models with new knowledge and skills while preserving previous capabilities~\citep{shi2025continual,wu2024continual}. Sequential LLM memory shares this temporal structure, but differs in a crucial way: most memory-augmented agents do not update model parameters, and instead rely on external, prompt-mediated, or retrieval-based memory states~\citep{zheng2023synapse,liang2025sage,li2025memos,wei2025evo}. Therefore, continual learning metrics cannot be directly reused without adaptation. We adapt their diagnostic principles to sequential LLM memory by defining online utility, hold-out generalization, backward transfer, and forgetting over evolving memory states, enabling fine-grained analysis of memory behavior beyond aggregate performance.

\section{\textsc{SeqMem-Eval}: A Diagnostic Evaluation Framework}

We propose \textsc{SeqMem-Eval}, a diagnostic framework for evaluating sequentially evolving LLM memory. Unlike classical continual learning, this setting keeps LLMs fixed and changes behavior through external memory updates, retrieval, and prompt construction. Thus, evaluation should go beyond endpoint performance to measure whether memory is useful online, generalizes to unseen tasks, consolidates past experience, preserves acquired utility, and remains computationally efficient.

\subsection{Sequential Memory Evaluation Setting}

Following Wei et al.~\cite{wei2025evo}, we consider a sequentially evolving memory setting, where an LLM $\gL$ interacts with a stream of tasks and maintains an external memory that is updated over time. 
Let $\mathcal{D}=\{(x_t,y_t)\}_{t=1}^{T}$ denote a task sequence, where $x_t$ is the input at step $t$ and $y_t$ is the corresponding target. 
At each step, the model maintains a memory state $M_t$, which may contain raw trajectories, retrieved examples, summaries, workflows, reflections, or other forms of accumulated experience from previous interactions.
Given input $x_t$, the system retrieves or constructs a context $C_t$ from the current memory state $M_t$, and the LLM produces a prediction $\hat{y}_t = \gL(x_t, C_t)$.
After prediction, the system may receive feedback $f_t$, such as correctness signals, execution results, or environment feedback. 
The memory is then updated by a method-specific update function:
\begin{equation}
M_{t+1} = \texttt{Update}(M_t, x_t, \hat{y}_t, f_t; \gL),
\end{equation}
where $\gL$ is included because some memory methods use the LLM itself to generate, refine, compress, or reorganize memory entries. 
For methods that do not rely on LLM-based memory updates, this argument can be omitted, and $\texttt{Update}$ reduces to a non-parametric operation, such as appending the current trajectory or updating a retrieval index.

This formulation abstracts a broad class of sequential memory methods, including retrieval-based memory, summarization-based memory, workflow memory, and reflection-based memory. We define diagnostic metrics over the evolving memory states $\{M_t\}_{t=1}^{T}$ to evaluate different aspects of memory behavior. Before introducing the formal definitions in the following subsections, Table~\ref{tab:seqmem_metrics_summary} provides an overview of the \textsc{SeqMem-Eval} metrics and the memory behavior captured by each metric.

\begin{table}[t]
\centering
\caption{
Overview of \textsc{SeqMem-Eval} diagnostics. 
The metrics cover five complementary perspectives of sequential memory evaluation, with formal definitions provided in the following subsections.
}
\label{tab:seqmem_metrics_summary}
\small
\setlength{\tabcolsep}{3.5pt}
\begin{tabular}{lll}
\toprule
Perspective & Metric & What it measures \\
\midrule

\multirow{4}{*}{Online Utility}
& OnlineAcc & Final cumulative online performance. \\
& PED & Peak-to-end degradation in the online trajectory. \\
& MER & Recovery from the worst online phase. \\
& $r_{\min}$ & Timing of the lowest online performance. \\

\midrule

\multirow{2}{*}{Hold-out Generalization}
& HoldOutAcc & Final performance on unseen examples. \\
& $\mathrm{Trend}_{\mathrm{HO}}$ & Direction of hold-out performance over memory evolution. \\

\midrule

\multirow{2}{*}{Backward Transfer}
& $\mathrm{BWT}(t)$ & Effect of later memory updates on earlier examples. \\
& IV & Immediate reusability of the newly updated memory. \\

\midrule

Forgetting
& $\mathrm{F}(t)$ & Loss of previously achieved capability over time. \\

\midrule

\multirow{2}{*}{Efficiency}
& Token Consumption & Total token cost of sequential evaluation. \\
& Runtime & End-to-end wall-clock time. \\

\bottomrule
\end{tabular}
\end{table}

\subsection{Online Utility}

\emph{Online utility} reflects how well the model performs as it progressively updates its memory over the task sequence. 
Rather than reducing online performance to a single scalar, we treat it as a trajectory that captures how the model's behavior evolves over time.

Formally, let $\mathcal{D}=\{(x_\tau,y_\tau)\}_{\tau=1}^{T}$ denote the online task sequence. 
For each step $\tau$, we record the per-step online performance and the cumulative online accuracy as
\begin{equation}
A(\tau)=\mathrm{Acc}(x_\tau; M_\tau),
\quad
\overline{A}(\tau)=\frac{1}{\tau}\sum_{i=1}^{\tau}A(i),
\end{equation}
where $\mathrm{Acc}(x;M)\in\{0,1\}$ indicates whether the model correctly answers $x$ using memory state $M$.

We report the following online utility metrics:

\noindent \textbullet\ \textbf{Online Accuracy (OnlineAcc).}
    The final value of the cumulative online accuracy curve:
    \begin{equation}
    \mathrm{OnlineAcc}
    =
    \overline{A}(T).
    \end{equation}
    While $\mathrm{OnlineAcc}$ summarizes endpoint performance, it does not capture how the trajectory reaches that endpoint. We further derive three trajectory-level diagnostics from the cumulative curve.

\noindent\textbullet\  \textbf{Peak-to-End Drop (PED).}
    The drop from the best cumulative online performance to the final value:
    \begin{equation}
    \mathrm{PED}
    =
    \max_{\tau\in[1,T]}\overline{A}(\tau)-\overline{A}(T).
    \end{equation}
    A larger $\mathrm{PED}$ indicates that the method reaches strong intermediate performance but later degrades.

\noindent\textbullet\  \textbf{Minimum-to-End Recovery (MER).}
    The recovery from the worst cumulative online performance to the final value:
    \begin{equation}
    \mathrm{MER}
    =
    \overline{A}(T)-\min_{\tau\in[1,T]}\overline{A}(\tau).
    \end{equation}
    A larger $\mathrm{MER}$ indicates that the method improves after a low-performance phase.

\noindent\textbullet\  \textbf{Extremum Timing $r_{\min}$ [Optional].}
    The relative location of the minimum cumulative online accuracy:
    \begin{equation}
    r_{\min}
    =
    \frac{1}{T}\arg\min_{\tau\in[1,T]}\overline{A}(\tau).
    \end{equation}
    This secondary diagnostic helps distinguish trajectory shapes with similar $\mathrm{MER}$ and $\mathrm{PED}$, such as gradual improvement versus delayed recovery.

Overall, these diagnostics complement $\mathrm{OnlineAcc}$ by characterizing whether the online trajectory reflects improvement, recovery, degradation, or volatile memory dynamics. 
Table~\ref{tab:trajectory_patterns} summarizes the qualitative interpretation of these trajectory patterns. 
Since strong online performance may still arise from short-term adaptation or task-order effects, we next evaluate whether the accumulated memory generalizes beyond the observed stream using hold-out test sets.

\begin{table}[t]
\centering
\caption{
Qualitative interpretation of online trajectory diagnostics. 
$\mathrm{MER}$ and $\mathrm{PED}$ provide coarse indicators of desirable or undesirable online dynamics, while $r_{\min}$ further refines the interpretation by distinguishing different improvement patterns.
}
\label{tab:trajectory_patterns}
\small
\begin{tabular}{lllll}
\toprule
Trajectory pattern & MER & PED & $r_{\min}$ & Preference (Interpretation) \\
\midrule
Gradual improvement 
& High & Low / 0 & Early & \cmark\ Preferred (Effective accumulation) \\
Drop-then-recover 
& High & Low / 0 & Middle & \cmark\ Acceptable (Delayed recovery) \\
Early peak then degradation 
& Low / 0 & High & Late & \xmark\ Undesirable (Unstable evolution) \\
Rapid drop then stabilization 
& Low / 0 & High & Late & \xmark\ Undesirable (Persistent degradation) \\
Stable but non-improving 
& Low & Low & Arbitrary & \nmark\ Mixed (Limited memory effect) \\
Highly fluctuating 
& High & High & Arbitrary & \nmark\ Mixed (Volatile memory dynamics) \\
\bottomrule
\end{tabular}
\vspace{-0.2in}
\end{table}

\subsection{Hold-out Generalization}

Online utility measures performance within the observed task stream, but strong online performance does not necessarily imply that memory has learned reusable knowledge. 
To evaluate this aspect, we measure \emph{hold-out generalization} on a fixed set of unseen tasks under different memory states.

Let $\mathcal{D}_{\mathrm{test}}=\{x_i^{\mathrm{test}}\}_{i=1}^{M}$ denote a hold-out test set that is never used for memory updates. 
At step $\tau$, after processing the first $\tau$ online tasks, the model maintains memory state $M_\tau$.

\noindent\textbullet\ \textbf{Hold-out Accuracy (HoldOutAcc).}
We define the hold-out accuracy trajectory and its final value as
\begin{equation}
H(\tau)
=
\frac{1}{M}
\sum_{i=1}^{M}
\mathrm{Acc}(x_i^{\mathrm{test}}; M_\tau),
\quad
\mathrm{HoldOutAcc}=H(T).
\end{equation}
The final value $\mathrm{HoldOutAcc}$ measures the generalization ability of the final memory state after processing the full online sequence.

\noindent\textbullet\ \textbf{Hold-out Trend ($\mathrm{Trend}_{\mathrm{HO}}$).}
To capture the overall direction of memory generalization, we compute the slope of the least-squares line fitted to the normalized hold-out trajectory:
\begin{equation}
\mathrm{Trend}_{\mathrm{HO}}
=
\mathrm{slope}\left(\{(\tau/T, H(\tau))\}_{\tau=1}^{T}\right).
\end{equation}
A positive $\mathrm{Trend}_{\mathrm{HO}}$ indicates that hold-out performance generally improves as more experiences are incorporated into memory, while a flat or negative value suggests limited generalization benefit or possible degradation.

\subsection{Backward Transfer}

Hold-out generalization evaluates whether memory improves performance on unseen tasks. 
We next consider the complementary question: how later memory updates affect previously encountered tasks. 
This is related to \textit{backward transfer} in continual learning~\citep{lopez2017gradient}, but here it is defined over evolving external memory states rather than model parameters.

\noindent\textbullet\ \textbf{Backward Transfer (BWT).}
For a temporal horizon $t$, we define
\begin{equation}
\mathrm{BWT}(t)
=
\frac{1}{T-t}
\sum_{\tau=1}^{T-t}
\left[
\mathrm{Acc}(x_{\tau}; M_{\tau+t})
-
\mathrm{Acc}(x_{\tau}; M_{\tau})
\right],
\end{equation}
where $M_{\tau}$ denotes the memory state after processing the $\tau$-th task, and $M_{\tau+t}$ denotes the memory state after the next $t$ updates. 
A positive $\mathrm{BWT}(t)$ indicates that later memory updates improve performance on earlier tasks, a value close to zero suggests limited retrospective effect, and a negative value indicates negative backward transfer.

\noindent\textbullet\ \textbf{Immediate Validity (IV).}
We define \emph{Immediate Validity} as the one-step backward transfer:
\begin{equation}
\mathrm{IV}
=
\mathrm{BWT}(1)
=
\frac{1}{T-1}
\sum_{\tau=1}^{T-1}
\left[
\mathrm{Acc}(x_{\tau}; M_{\tau+1})
-
\mathrm{Acc}(x_{\tau}; M_{\tau})
\right].
\end{equation}
$\mathrm{IV}$ evaluates whether the memory update induced by the current task immediately makes that experience reusable. 
A positive $\mathrm{IV}$ suggests that the update effectively incorporates the new experience, while a value close to zero or below indicates that the update provides limited reusable benefit.

\subsection{Forgetting}

BWT measures net transfer from later memory updates to earlier tasks, but it does not capture whether a capability that was temporarily acquired is later lost. 
We therefore measure \emph{forgetting}, which evaluates whether previously achieved performance is retained over time.

\noindent\textbullet\ \textbf{Forgetting.}
For a temporal horizon $t$, we define forgetting as
\begin{equation}
\mathrm{F}(t)
=
\frac{1}{T-t}
\sum_{\tau=1}^{T-t}
\left[
\max_{\tau' \in [\tau,\,\tau+t]}
\mathrm{Acc}(x_{\tau}; M_{\tau'})
-
\mathrm{Acc}(x_{\tau}; M_{\tau+t})
\right],
\end{equation}
where the first term is the best performance achieved on $x_\tau$ within $[\tau,\tau+t]$, and the second term is its performance after $t$ additional memory updates. 
A larger $\mathrm{F}(t)$ indicates stronger loss of previously acquired capability, while values close to zero suggest stable retention.

Computing $\mathrm{F}(t)$ exactly requires evaluating each past task under all intermediate memory states, which can be expensive for long sequences. 
In practice, we use a checkpoint-based approximation that reuses the evaluations performed for backward transfer. 
Let $\mathcal{T}=\{t_1,t_2,\dots\}$ denote a discrete set of temporal horizons. 
For each task $x_\tau$ and horizon $t$, we approximate
\begin{equation}
\max_{\tau' \in [\tau,\,\tau+t]}
\mathrm{Acc}(x_{\tau}; M_{\tau'})
\approx
\max_{t_i \in \mathcal{T},\, t_i \le t}
\mathrm{Acc}(x_{\tau}; M_{\tau+t_i}).
\end{equation}
This approximation reduces computation while still capturing whether performance achieved at earlier checkpoints is lost after subsequent memory updates. 
Together, BWT and forgetting provide complementary views: BWT measures net retrospective transfer, whereas forgetting measures degradation from the best previously achieved performance.

\subsection{Efficiency}

Memory methods may improve performance by using longer contexts, storing more experiences, or invoking the LLM multiple times. 
Thus, memory quality should be evaluated together with computational cost.
We measure efficiency along two dimensions.

\noindent$\bullet$ \textbf{Token Consumption.}
We report the total number of tokens used during sequential evaluation, including prompts, retrieved memories, final answers, and intermediate reasoning or memory updates.

\noindent$\bullet$ \textbf{Runtime.}
We report the wall-clock time required to process the full task sequence, including LLM calls and overhead from retrieval, memory construction, and summarization.

Together, these metrics reveal whether performance gains come from better memory behavior or substantially larger computational budgets.

\section{Experiments and Analysis}
We evaluate sequentially evolving LLM memory under the proposed \textsc{SeqMem-Eval} framework across diverse tasks, models, and representative memory methods. We provide the detailed experimental setup in Appendix~\ref{appx:exp-setup}. 
Rather than only comparing final accuracy, we organize our analysis around the following research questions (RQs):
\begin{itemize}[leftmargin=*]
    \item \textbf{RQ1: Online Utility.} How does memory affect performance during sequential inference, and what online trajectory patterns emerge as memory evolves?
    
    \item \textbf{RQ2: Hold-out Generalization.} 
    Does the evolving memory improve performance on unseen tasks, including in-distribution and out-of-distribution hold-out sets?
    
    \item \textbf{RQ3: Backward Transfer.} 
    Do later memory updates help consolidate knowledge for previously encountered tasks, or do they induce negative transfer?
    
    \item \textbf{RQ4: Forgetting.} 
    Does the memory retain capabilities achieved during the sequence, or does performance on earlier tasks degrade over time?

    \item \textbf{RQ5: Efficiency.} What are the efficiency trade-offs of different memory mechanisms in terms of computational cost?
\end{itemize}


\subsection{Analysis of Online Utility (RQ1)}
\label{sec:rq1-online}

\begin{figure}[H]
    \centering
    \begin{subfigure}{\textwidth}
        \centering
        \includegraphics[width=\textwidth]{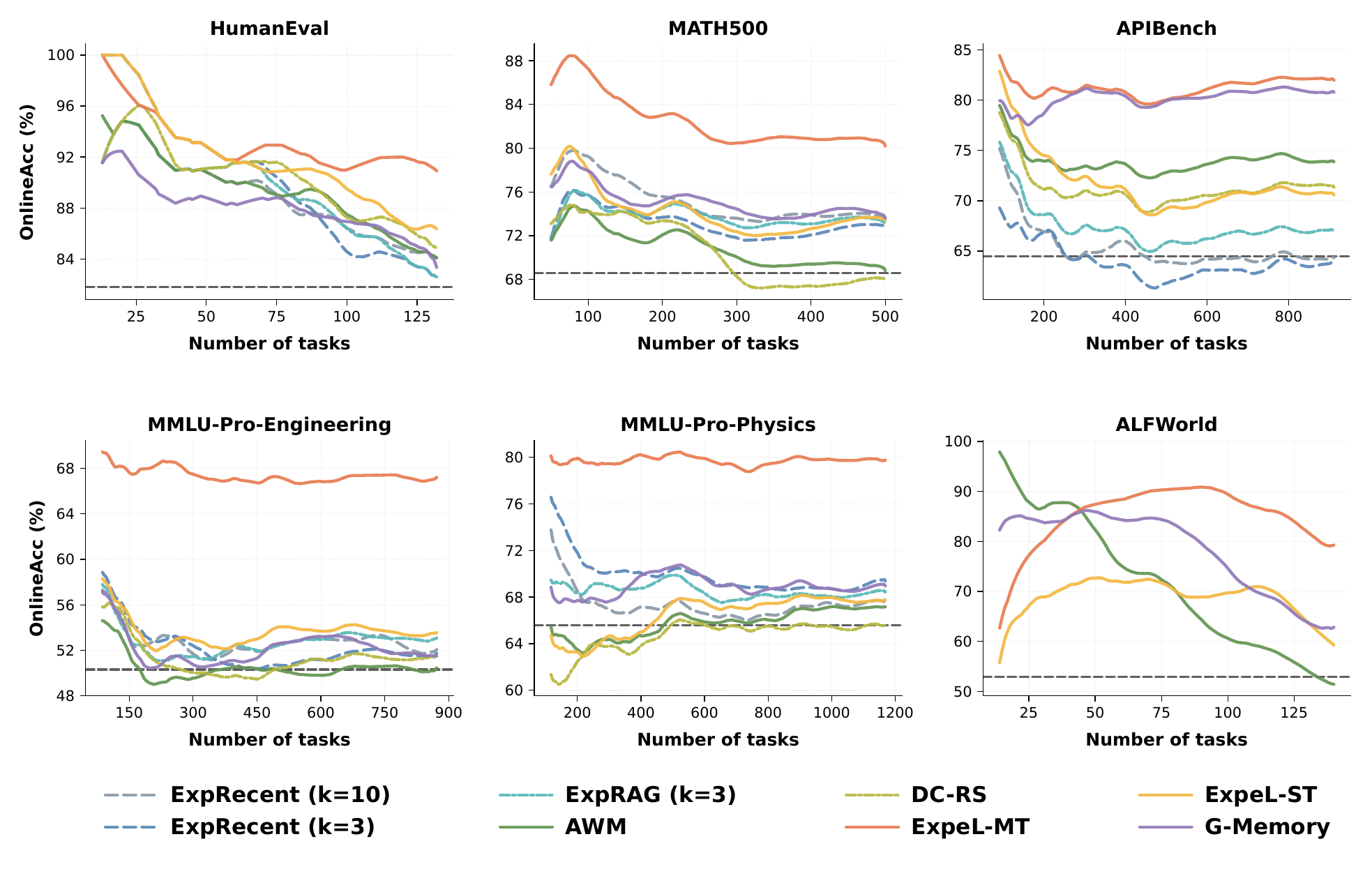}
        \caption{Qwen3-8B}
    \end{subfigure}


    \begin{subfigure}{\textwidth}
        \centering
        \includegraphics[width=\textwidth]{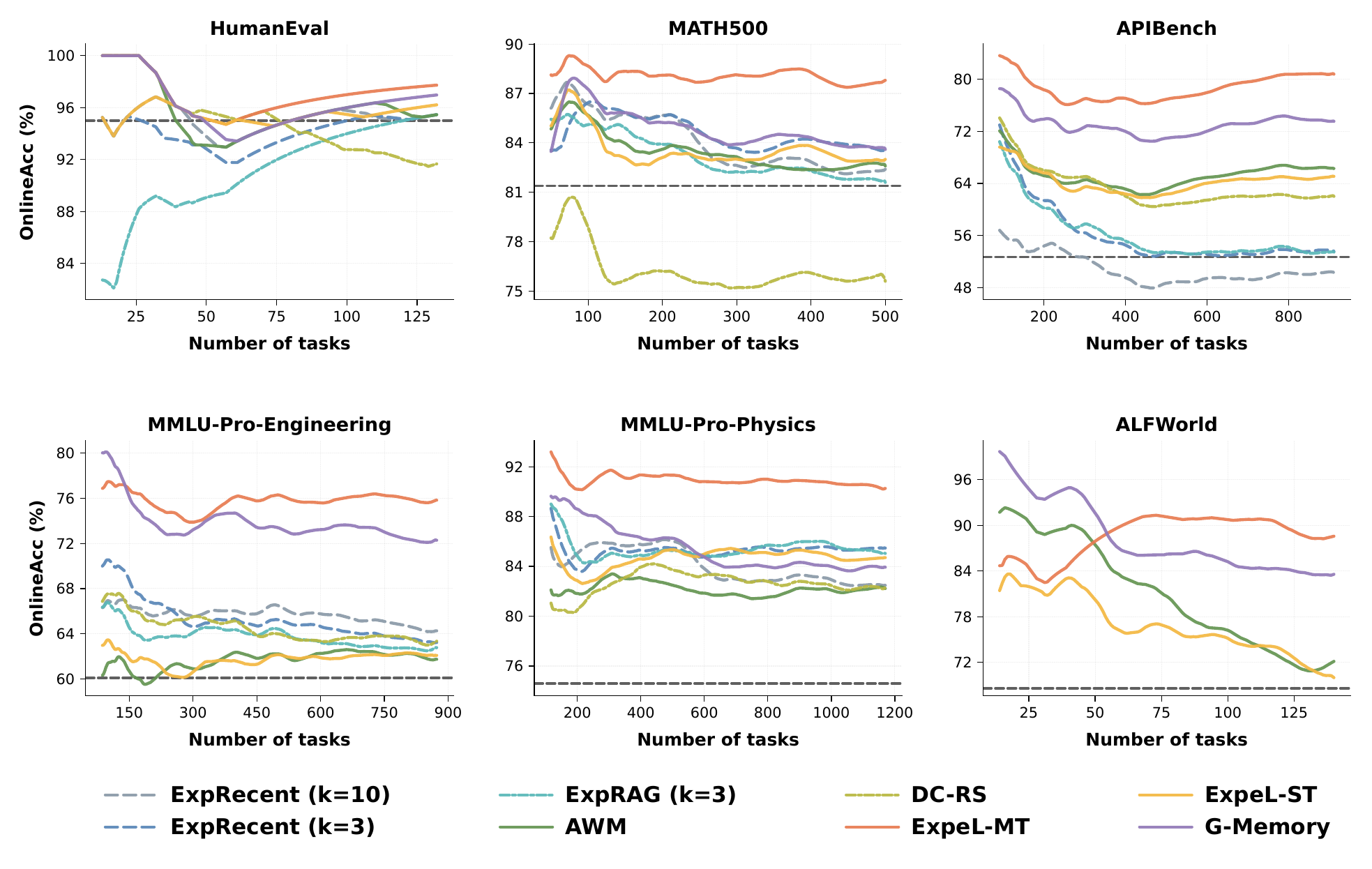}
        \caption{MiniMax-M2.7}
        \label{fig:minimax_online_acc}
    \end{subfigure}
    \caption{Online accuracy over sequential steps for different models.}
     \label{fig:qwen_minimax_online}
\end{figure}

%
\textbf{\textsc{Finding 1.} Memory improves aggregate online accuracy, but the gains do not imply stable accumulation.}
Table~\ref{tab:online_acc} reports the standard aggregate view of online performance. Under this view, memory-augmented methods often appear beneficial: most methods improve over the memory-free baseline across datasets, and stronger memory mechanisms such as G-Memory and ExpeL-MT achieve clear gains across both LLM backbones.
However, the cumulative trajectories in Figure~\ref{fig:qwen_minimax_online} reveal a more nuanced picture. 
We also observe that \textit{monotonic improvement is largely absent}: across HumanEval and MMLU-Pro-Engineering, many methods either decline after an initially strong phase or rapidly drop and then plateau. This suggests that additional memory updates do not consistently translate into sustained experience accumulation.

\begin{table*}[ht]
\centering
\small
\setlength{\tabcolsep}{4pt}
\caption{Accuracy (\%) across LLMs, tasks, and methods. Colored subscripts indicate changes relative to the corresponding memory-free baseline. Best results are bolded.}
\label{tab:online_acc}
\resizebox{\textwidth}{!}{
\begin{tabular}{llcccccc}
\toprule
\textbf{Model} & \textbf{Method}
& \textbf{HumanEval}
& \textbf{MATH500}
& \textbf{APIBench}
& \textbf{MMLU-Eng.}
& \textbf{MMLU-Phys.}
& \textbf{ALFWorld} \\
\midrule

\multirow{9}{*}{Qwen3-8B}
& Memory-free & 81.8 & 68.6 & 64.5 & 50.3 & 65.6 & 52.9 \\
& ExpRec$_{k=10}$ & 84.1\up{2.3} & 73.6\up{5.0} & 64.3\down{0.2} & 52.1\up{1.8} & 67.5\up{1.9} & -- \\
& ExpRec$_{k=3}$ & 82.6\up{0.8} & 72.8\up{4.2} & 63.9\down{0.6} & 51.7\up{1.4} & 69.4\up{3.8} & -- \\
& ExpRAG$_{k=3}$ & 82.6\up{0.8} & 73.0\up{4.4} & 67.0\up{2.5} & 53.1\up{2.8} & 68.4\up{2.8} & -- \\
& DC-RS & 84.8\up{3.0} & 68.0\down{0.6} & 71.4\up{6.9} & 51.5\up{1.2} & 65.5\down{0.1} & -- \\
& AWM & 84.1\up{2.3} & 68.8\up{0.2} & 73.9\up{9.4} & 50.5\up{0.2} & 67.2\up{1.6} & 51.4\down{1.5} \\
& G-Memory & 83.3\up{1.5} & 73.6\up{5.0} & 80.8\up{16.3} & 51.5\up{1.2} & 68.9\up{3.3} & 62.9\up{10.0} \\
& ExpeL-ST & 86.4\up{4.6} & 73.4\up{4.8} & 70.6\up{6.1} & 53.6\up{3.3} & 67.8\up{2.2} & 59.3\up{6.4} \\
& ExpeL-MT & \textbf{90.9}\up{9.1} & \textbf{80.2}\up{11.6} & \textbf{82.0}\up{17.5} & \textbf{67.2}\up{16.9} & \textbf{79.7}\up{14.1} & \textbf{79.3}\up{26.4} \\

\midrule

\multirow{9}{*}{MiniMax-M2.7}
& Memory-free & 95.0 & 81.4 & 52.7 & 60.1 & 74.6 & 68.6 \\
& ExpRec$_{k=10}$ & 95.5\up{0.5} & 82.4\up{1.0} & 50.3\down{2.4} & 64.2\up{4.1} & 82.5\up{7.9} & -- \\
& ExpRec$_{k=3}$ & 95.5\up{0.5} & 83.6\up{2.2} & 53.6\up{0.9} & 63.2\up{3.1} & 85.5\up{10.9} & -- \\
& ExpRAG$_{k=3}$ & 95.5\up{0.5} & 81.6\up{0.2} & 53.5\up{0.8} & 62.7\up{2.6} & 85.0\up{10.4} & -- \\
& DC-RS & 91.7\down{3.3} & 75.6\down{5.8} & 62.0\up{9.3} & 63.3\up{3.2} & 82.3\up{7.7} & -- \\
& AWM & 95.5\up{0.5} & 82.6\up{1.2} & 66.3\up{13.6} & 61.7\up{1.6} & 82.2\up{7.6} & 72.1\up{3.5} \\
& G-Memory & 97.0\up{2.0} & 83.6\up{2.2} & 73.5\up{20.8} & 72.2\up{12.1} & 83.9\up{9.3} & 83.6\up{15.0} \\
& ExpeL-ST & 96.2\up{1.2} & 83.0\up{1.6} & 65.1\up{12.4} & 62.0\up{1.9} & 84.7\up{10.1} & 70.0\up{1.4} \\
& ExpeL-MT & \textbf{97.7}\up{2.7} & \textbf{87.8}\up{6.4} & \textbf{80.8}\up{28.1} & \textbf{75.8}\up{15.7} & \textbf{90.2}\up{15.6} & \textbf{88.6}\up{20.0} \\

\bottomrule
\end{tabular}
}
\end{table*}

\textbf{\textsc{Finding 2.} Methods with similar online accuracy can exhibit substantially different memory dynamics.}
Table~\ref{tab:online_dynamics} summarizes the MER and PED diagnostics across all datasets and methods, together with the extremum timing metric $r_{\min}$. The results show that methods with comparable final \textsc{OnlineAcc} can still occupy substantially different regions in the diagnostic space. On HumanEval, several methods have non-trivial PED but small MER, corresponding to early high performance followed by degradation. On ALFWorld, the separation is more pronounced: ExpeL-MT achieves high MER with relatively lower PED compared with other agentic methods, while AWM-Online suffers from large PED and near-zero MER, indicating severe peak-to-end degradation without recovery.

Overall, \textbf{RQ1} shows that aggregate OnlineAcc provides only a coarse endpoint measure. Although memory methods often improve online performance on average, their trajectories reveal whether these gains are stable, recoverable, or transient. This supports the need to evaluate online utility as an evolving trajectory rather than a single aggregate score.

\begin{table*}[ht]
\centering
\small
\setlength{\tabcolsep}{4pt}
\caption{
Online trajectory diagnostics on Qwen3-8B and MiniMax-M2.7. 
Higher MER and lower PED are generally preferred; $r_{\min}$ indicates the relative timing of the lowest online performance and should be interpreted together with MER and PED.
}
\label{tab:online_dynamics}
\resizebox{\textwidth}{!}{
\begin{tabular}{lcccccccccccccccccc}
\toprule
\multirow{2}{*}{Method}
& \multicolumn{3}{c}{HumanEval}
& \multicolumn{3}{c}{MATH500}
& \multicolumn{3}{c}{APIBench}
& \multicolumn{3}{c}{MMLU-Eng.}
& \multicolumn{3}{c}{MMLU-Phys.}
& \multicolumn{3}{c}{ALFWorld} \\
\cmidrule(lr){2-4}
\cmidrule(lr){5-7}
\cmidrule(lr){8-10}
\cmidrule(lr){11-13}
\cmidrule(lr){14-16}
\cmidrule(lr){17-19}
& MER $\uparrow$ & PED $\downarrow$ & $r_{\min}\downarrow$
& MER $\uparrow$ & PED $\downarrow$ & $r_{\min}\downarrow$
& MER $\uparrow$ & PED $\downarrow$ & $r_{\min}\downarrow$
& MER $\uparrow$ & PED $\downarrow$ & $r_{\min}\downarrow$
& MER $\uparrow$ & PED $\downarrow$ & $r_{\min}\downarrow$
& MER $\uparrow$ & PED $\downarrow$ & $r_{\min}\downarrow$ \\
\midrule
\multicolumn{19}{c}{\textbf{Backbone LLM: Qwen3-8B}}\\ \midrule

\textbf{ExpRec$_{k=10}$}
& 0.001 & 0.121 & 0.99
& 0.007 & 0.084 & 0.66
& 0.019 & 0.113 & 0.28
& 0.013 & 0.067 & 0.38
& 0.018 & 0.060 & 0.61
& -- & -- & -- \\

ExpRec$_{k=3}$
& 0.001 & 0.174 & 0.99
& 0.018 & \textbf{0.061} & 0.62
& 0.034 & 0.070 & 0.50
& 0.020 & 0.088 & 0.50
& 0.011 & 0.076 & 0.89
& -- & -- & -- \\

ExpRAG$_{k=3}$
& 0.001 & 0.136 & 0.99
& \textbf{0.030} & \textbf{0.061} & \textbf{0.10}
& 0.031 & 0.095 & 0.50
& \textbf{0.033} & 0.055 & 0.24
& 0.018 & 0.024 & 0.17
& -- & -- & -- \\

DC-RS
& 0.001 & 0.120 & 0.99
& 0.013 & 0.087 & 0.62
& 0.036 & 0.076 & 0.50
& 0.028 & 0.073 & 0.50
& \textbf{0.072} & 0.010 & 0.12
& -- & -- & -- \\

AWM
& 0.001 & 0.121 & 0.99
& 0.000 & 0.084 & 1.00
& 0.024 & 0.060 & 0.50
& 0.031 & 0.057 & 0.24
& 0.047 & \textbf{0.003} & 0.18
& 0.000 & 0.486 & 1.00 \\

G-Memory
& 0.000 & 0.114 & 1.00
& 0.004 & 0.071 & 0.73
& \textbf{0.045} & \textbf{0.008} & \textbf{0.11}
& 0.022 & 0.068 & 0.23
& 0.034 & 0.022 & \textbf{0.10}
& 0.007 & 0.246 & 0.94 \\

ExpeL-ST
& \textbf{0.007} & 0.136 & 0.90
& 0.022 & 0.084 & 0.62
& 0.030 & 0.118 & 0.50
& 0.028 & 0.051 & 0.24
& 0.057 & \textbf{0.008} & 0.12
& 0.005 & 0.152 & 0.12 \\

ExpeL-MT
& 0.005 & \textbf{0.091} & \textbf{0.71}
& 0.003 & 0.100 & 0.62
& 0.035 & 0.013 & 0.19
& 0.013 & \textbf{0.035} & \textbf{0.16}
& 0.020 & 0.012 & 0.12
& \textbf{0.150} & \textbf{0.122} & \textbf{0.10} \\

\midrule
\multicolumn{19}{c}{\textbf{Backbone LLM: MiniMax-M2.7}}\\ \midrule


ExpRec$_{k=10}$
& 0.033 & 0.019 & 0.39
& 0.006 & 0.070 & 0.87
& 0.033 & 0.089 & 0.50
& 0.009 & 0.052 & \textbf{0.10}
& 0.002 & 0.039 & 0.92
& -- & -- & -- \\

ExpRec$_{k=3}$
& 0.046 & 0.010 & 0.42
& \textbf{0.028} & 0.047 & \textbf{0.10}
& 0.018 & 0.186 & 0.50
& 0.001 & 0.103 & 0.95
& 0.027 & 0.035 & 0.17
& -- & -- & -- \\

ExpRAG$_{k=3}$
& \textbf{0.185} & \textbf{0.000} & \textbf{0.10}
& 0.000 & 0.056 & 0.87
& 0.008 & 0.170 & 0.50
& 0.005 & 0.065 & 0.97
& 0.013 & 0.039 & 0.17
& -- & -- & -- \\

DC-RS
& 0.004 & 0.083 & 0.95
& 0.014 & 0.074 & 0.26
& 0.024 & 0.121 & 0.50
& 0.006 & 0.073 & 0.97
& 0.027 & 0.023 & 0.14
& -- & -- & -- \\

AWM
& 0.033 & 0.046 & 0.39
& 0.006 & 0.057 & 0.87
& 0.047 & 0.067 & 0.50
& \textbf{0.040} & \textbf{0.026} & \textbf{0.10}
& 0.019 & 0.018 & \textbf{0.11}
& 0.021 & 0.229 & 0.91 \\

G-Memory
& 0.042 & 0.030 & 0.42
& 0.002 & 0.066 & 0.87
& 0.040 & 0.070 & 0.50
& 0.004 & 0.091 & 0.97
& 0.004 & 0.066 & 0.89
& 0.004 & 0.164 & 0.94 \\

ExpeL-ST
& 0.039 & 0.012 & \textbf{0.10}
& 0.014 & 0.064 & 0.32
& 0.040 & 0.065 & 0.50
& 0.028 & 0.034 & 0.30
& \textbf{0.028} & \textbf{0.017} & 0.18
& 0.000 & 0.170 & 1.00 \\

ExpeL-MT
& 0.036 & 0.023 & 0.39
& 0.018 & \textbf{0.037} & 0.21
& \textbf{0.057} & \textbf{0.041} & \textbf{0.28}
& 0.025 & 0.040 & 0.34
& 0.009 & 0.030 & 0.17
& \textbf{0.086} & \textbf{0.034} & \textbf{0.18} \\

\bottomrule
\end{tabular}
}
\end{table*}


\subsection{Analysis of Hold-out Generalization (RQ2)}

\textbf{\textsc{Finding 3.} Online gains do not reliably translate to hold-out generalization.}
Table~\ref{tab:qwen_holdout} shows that final in-distribution (ID) hold-out generalization is often weak, even when memory methods improve online accuracy. Across several datasets, many methods only match or fall below the memory-free baseline on the final hold-out set. For example, on APIBench, most memory methods underperform the memory-free baseline in final hold-out accuracy, despite showing gains in aggregate online performance. This gap suggests that memory updates can improve performance on the observed stream without necessarily learning reusable knowledge that transfers to unseen examples.

\textbf{\textsc{Finding 4.} Final hold-out accuracy can hide unstable generalization dynamics.}
The trend values in Table~\ref{tab:qwen_holdout} show that endpoint hold-out accuracy provides an incomplete view of memory generalization. Positive $\mathrm{Trend}_{\mathrm{HO}}$ values are not consistent across methods or datasets, and several methods achieve competitive final hold-out accuracy despite negative trends. This indicates that their final memory state may not reflect stable improvement over time. The hold-out trajectories in Appendix~\ref{appx:id-holdout} further support this observation: intermediate memory states can outperform the final memory state, suggesting that useful memory may be overwritten, diluted, or destabilized by later updates. We provide a supporting case study in Appendix~\ref{appx:case-study}.

The OOD hold-out results in Appendix~\ref{appx:od-holdout-qwen} show a similar but stronger pattern, where cross-task generalization is substantially weaker than ID generalization and memory updates can even hurt performance compared with the memory-free baseline. Overall, \textbf{RQ2} shows that hold-out generalization should be analyzed through both endpoint performance and trajectory trends. These results highlight the need for memory methods to validate update quality, preserve useful intermediate states, and prevent harmful updates.

\begin{table}[t]
\centering
\caption{
In-distribution hold-out generalization on Qwen3-8B.We report the final $\mathrm{HoldOutAcc}$ using the final memory state and the hold-out trajectory trend $\mathrm{Trend}_{\mathrm{HO}}$. 
Higher values are better.
}
\label{tab:qwen_holdout}
\scriptsize
\setlength{\tabcolsep}{1pt}
\resizebox{\textwidth}{!}{
\begin{tabular}{lcccccccccccc}
\toprule
\multirow{2}{*}{\textbf{Method}}
& \multicolumn{2}{c}{\textbf{HumanEval}}
& \multicolumn{2}{c}{\textbf{MATH500}}
& \multicolumn{2}{c}{\textbf{APIBench}}
& \multicolumn{2}{c}{\textbf{MMLU-Eng}}
& \multicolumn{2}{c}{\textbf{MMLU-Phys}}
& \multicolumn{2}{c}{\textbf{ALFWorld}} \\

\cmidrule(lr){2-3} \cmidrule(lr){4-5} \cmidrule(lr){6-7}
\cmidrule(lr){8-9} \cmidrule(lr){10-11} \cmidrule(lr){12-13}

& HO-Acc & $\text{Trend}_{\text{HO}}$
& HO-Acc & $\text{Trend}_{\text{HO}}$
& HO-Acc & $\text{Trend}_{\text{HO}}$
& HO-Acc & $\text{Trend}_{\text{HO}}$
& HO-Acc & $\text{Trend}_{\text{HO}}$
& HO-Acc & $\text{Trend}_{\text{HO}}$\\

\midrule

Memory-free
& 75.0 & --
& 67.5 & --
& 67.8 & --
& 50.0 & --
& 69.0 & --
& 75.0 & -- \\

\midrule

ExpRec$_{k=10}$
& 78.1 & -0.001
& \textbf{71.4} & \textbf{+0.030}
& 54.2 & -0.111
& 53.1 & +0.039
& 66.7 & +0.004
& -- & -- \\

ExpRec$_{k=3}$
& 78.1 & -0.017
& 70.4 & -0.003
& 67.5 & -0.145
& 55.2 & +0.011
& 69.8 & +0.028
& -- & -- \\

ExpRAG
& 68.8 & -0.107
& 67.5 & -0.023
& 61.7 & -0.014
& 53.1 & -0.014
& 65.9 & +0.004
& -- & -- \\

DC-RS
& 75.0 & \textbf{+0.068}
& 61.8 & -0.010
& 67.5 & +0.015
& 51.0 & +0.006
& 65.9 & -0.004
& -- & -- \\

AWM
& 71.9 & -0.012
& 67.5 & +0.012
& 65.0 & -0.053
& 53.1 & \textbf{+0.089}
& 67.4 & -0.008
& 58.3 & +0.032 \\

ExpeL-ST
& \textbf{78.1} & -0.012
& 70.0 & -0.014
& 66.7 & -0.054
& \textbf{58.3} & +0.066
& 65.9 & -0.004
& 58.3 & +0.088 \\

ExpeL-MT
& 75.0 & -0.005
& 67.5 & +0.009
& \textbf{73.3} & \textbf{+0.072}
& 54.2 & -0.095
& \textbf{72.1} & \textbf{+0.057}
& 62.5 & \textbf{+0.220} \\

G-Memory
& 75.0 & -0.034
& 68.6 & -0.006
& 67.5 & -0.064
& 57.3 & +0.030
& 68.2 & -0.002
& \textbf{70.8} & -0.090 \\

\bottomrule
\end{tabular}
}
\end{table}

\subsection{Analysis of Backward Transfer (RQ3)}

\begin{figure}[t]
    \centering
    \includegraphics[width=\textwidth]{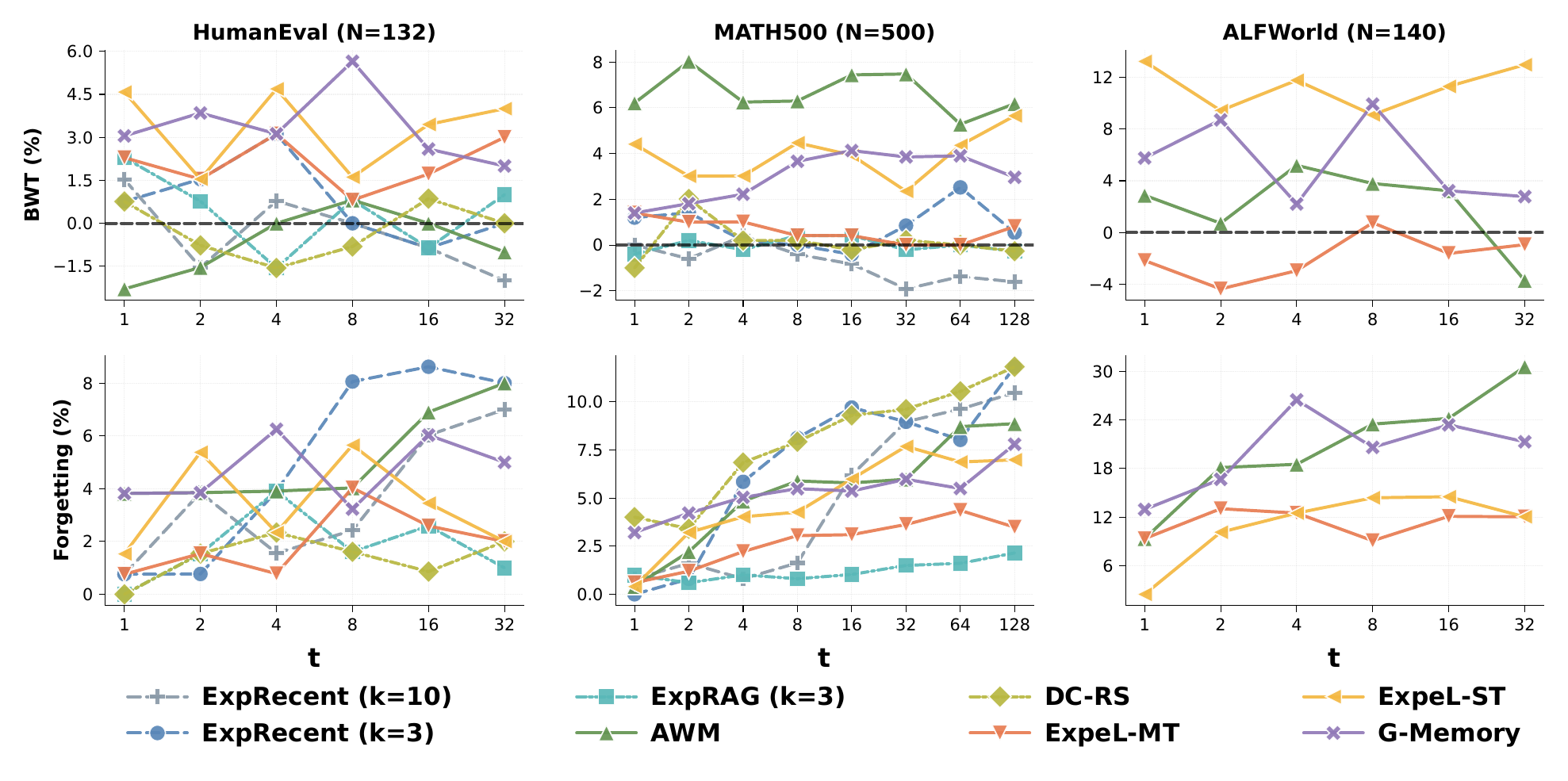}
    \caption{Backward transfer (top row) and forgetting (bottom row) for Qwen3-8B across different temporal horizons $t$. Higher BWT is better, while lower forgetting indicates better retention.}
    \label{fig:qwen_transfer}
\end{figure}

\textbf{\textsc{Finding 5.} Later memory updates provide short-term benefits, but do not reliably consolidate transferable knowledge.}
As shown in the top row of Figure~\ref{fig:qwen_transfer}, BWT is often positive at short horizons, indicating that recent memory updates can improve performance on earlier tasks. However, this effect does not consistently strengthen as the horizon $t$ grows. Instead, most methods exhibit fluctuating or diminishing BWT over longer horizons, suggesting that later memory updates do not reliably accumulate into increasingly useful retrospective knowledge.

This overall pattern arises from different failure modes across memory designs. Retrieval-based methods reveal a recency--retention trade-off: ExpRecent can exhibit negative BWT once past tasks fall outside its recent window, while ExpRAG is more stable but often remains near zero, indicating limited consolidation beyond instance retrieval. More structured memory methods, such as G-Memory, sometimes achieve stronger positive BWT, but their curves are still highly non-monotonic. These results suggest that current memory methods can reuse nearby or directly relevant experiences, but often fail to transform longer streams of interaction into stable transferable knowledge.

\textbf{\textsc{Finding 6.} Immediate validity is often weak even right after a memory update.}
The case $t=1$ corresponds to immediate validity, which measures whether the memory update induced by the current task improves performance on that same task. Surprisingly, although the memory has just incorporated the current task and its feedback, $\mathrm{BWT}(1)$ is often below $5\%$ and can even be negative. This suggests that many memory updates are relatively shallow: they may append, retrieve, or summarize the latest interaction, but do not necessarily extract reusable information that can immediately benefit the same task.

Overall, \textbf{RQ3} reveals a key limitation of existing memory mechanisms: they can often absorb local feedback, but they do not reliably consolidate it into transferable knowledge that benefits earlier tasks over longer horizons. This highlights the need for memory updates that go beyond storage and perform more effective reflection, abstraction, and validation.

\subsection{Analysis of Forgetting (RQ4)} 
\textbf{\textsc{Finding 7.} Sequential memory updates often interfere with previously acquired utility.}
The bottom row of Figure~\ref{fig:qwen_transfer} shows that forgetting generally increases with the horizon $t$. This indicates that performance previously achieved on earlier tasks is often not retained after additional memory updates. As more experiences are incorporated, memory states may overwrite useful information, dilute earlier knowledge, or retrieve less relevant content for past tasks.

Together with the BWT results, \textbf{RQ4} reveals a central tension in sequential LLM memory: methods can often incorporate recent feedback and sometimes improve past performance, but they struggle to preserve these gains over longer horizons. Future memory mechanisms should therefore not only add or summarize new experiences, but also preserve useful knowledge, validate updates, and prevent interference between old and new memory content.

\begin{wrapfigure}{r}{0.5\textwidth}
\vspace{-0.2in}
		\centering
\captionsetup[sub]{skip=-1pt}
\includegraphics[width=1\linewidth]{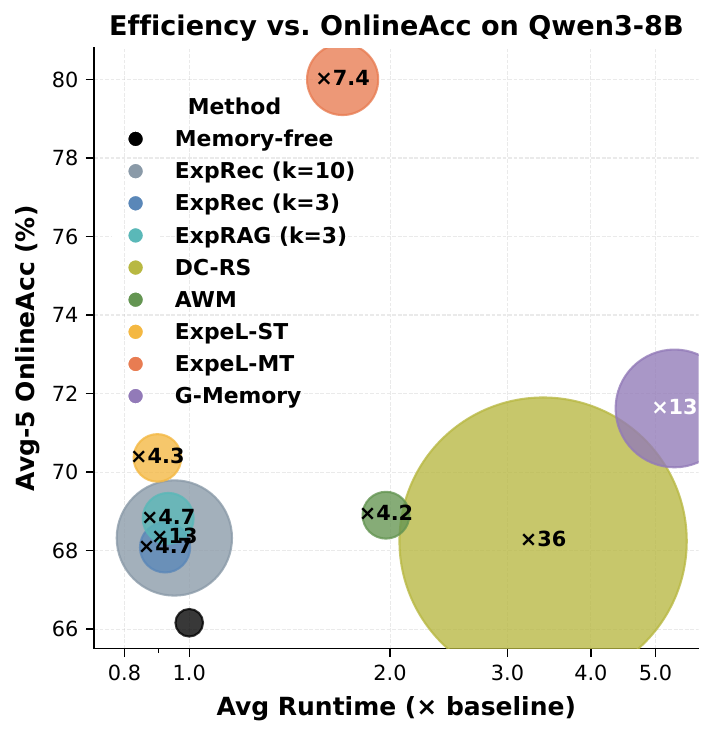}
\vspace{-0.2in}
\caption{
Efficiency--performance trade-off on Qwen3-8B. 
Bubble labels indicate total token usage normalized by the memory-free baseline.
}
\vspace{-0.2in}
\label{fig:efficiency_acc_tradeoff}
\end{wrapfigure}

\subsection{Analysis of Efficiency (RQ5)}

\textbf{\textsc{Finding 8.} Stronger memory performance often comes with substantial token and runtime overhead.}
Figure~\ref{fig:efficiency_acc_tradeoff} summarizes the efficiency--performance trade-off on Qwen3-8B. Most memory-augmented methods incur clear overhead over the memory-free baseline. Even lightweight retrieval methods such as ExpRecent and ExpRAG require about $4.3$--$4.7\times$ more tokens while providing only modest gains. More complex methods often achieve stronger performance but at substantially higher cost: ExpeL-MT obtains the best Avg-5 accuracy with $7.4\times$ token usage and the largest runtime increase, while G-Memory uses around $13\times$ more tokens. DC-RS is especially costly, requiring about $3.2\times$ runtime and $36\times$ token usage, yet performs close to much cheaper methods.

Overall, RQ5 shows that memory effectiveness should be interpreted together with computational cost. Strong performance can reflect substantially higher token and latency budgets, rather than more efficient memory utilization. Detailed runtime and token breakdowns are provided in Appendix~\ref{appx:efficiency}.

\section{Overall Comparison}
\label{overall_comparison}

\begin{wrapfigure}{r}{0.49\textwidth}
\vspace{-0.2in}
\centering
\captionsetup[sub]{skip=-1pt}
\includegraphics[width=1\linewidth]{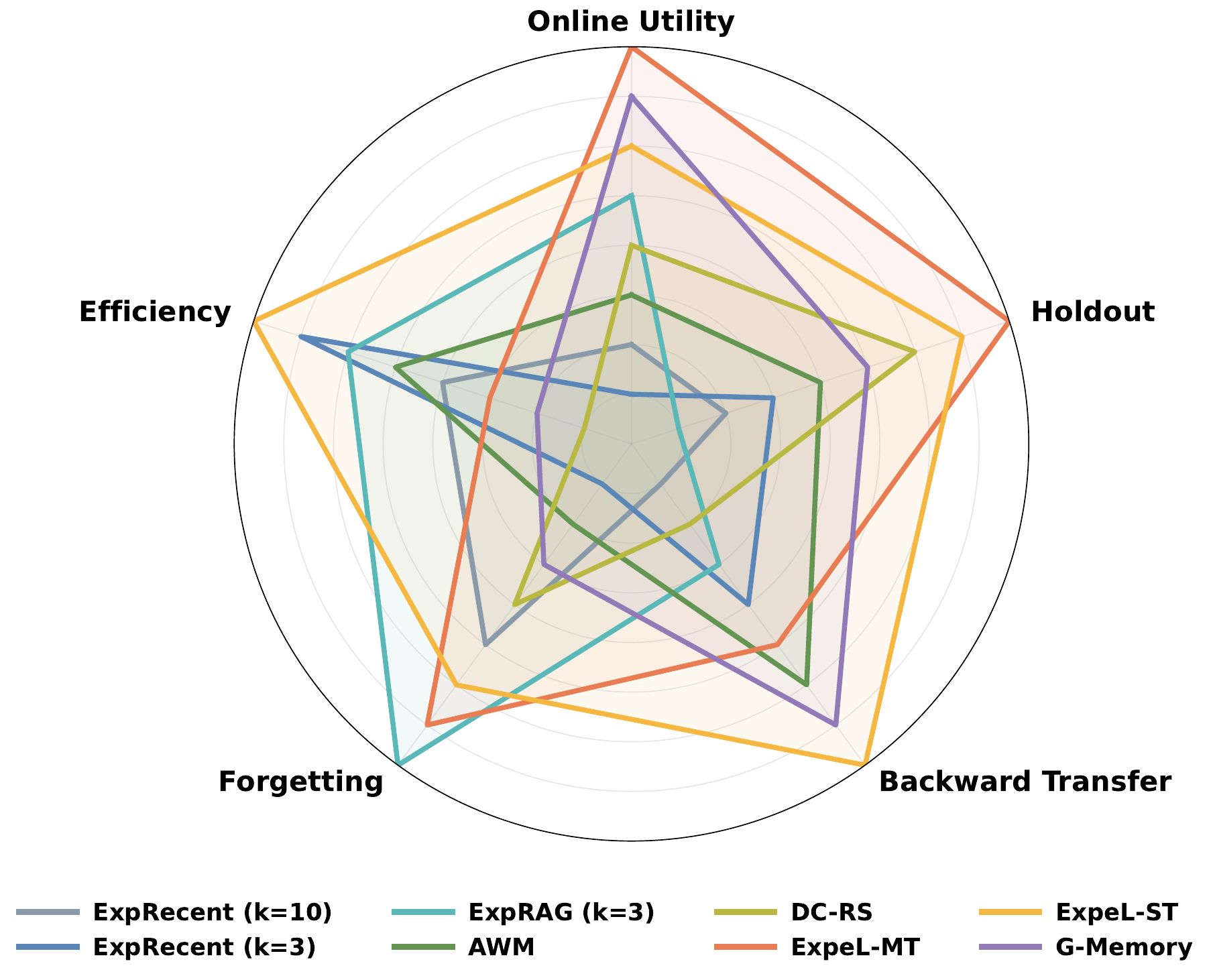}
\vspace{-0.2in}
\caption{
Overall radar comparison on Qwen3-8B across multiple dimensions. Larger radial values indicate better within-dimension rankings.
}
\vspace{-0.2in}
\label{fig:radar_qwen}
\end{wrapfigure}

To provide a holistic view of method behavior, Figure~\ref{fig:radar_qwen} summarizes representative memory mechanisms across the five diagnostic dimensions of \textsc{SeqMem-Eval}. For each dimension, we normalize the underlying metrics, average them into a dimension-level score, and rank methods accordingly. The radar plot reports these relative ranks to provide a compact profile of method trade-offs.

The radar profiles reveal that different algorithmic designs lead to distinct trade-offs. ExpeL-MT performs strongly on online utility and hold-out generalization, which is consistent with its use of multiple attempts and reflection-based experience extraction: repeated trials increase the chance of solving the current task, while distilled experiential rules can provide reusable signals for later tasks. However, the same design also introduces substantial generation and update overhead, explaining its weaker efficiency profile. ExpeL-ST removes the multi-try component and therefore offers a different trade-off: it retains some benefits of reflection-based memory while reducing the cost associated with repeated inference, suggesting that experience abstraction can be useful even without aggressive inference-time search.

Retrieval-based methods exhibit the opposite behavior. ExpRecent and ExpRAG are comparatively efficient because they mainly reuse stored examples through recent-window or similarity-based retrieval, without expensive memory synthesis. However, their weaker profiles on generalization and transfer suggest that instance-level retrieval alone provides limited abstraction: it can reuse nearby or similar experiences, but does not reliably consolidate them into higher-level knowledge that transfers across time or tasks. This observation is consistent with the BWT analysis, where retrieval-based methods show limited consolidation beyond local reuse.

Structured memory methods occupy an intermediate regime. G-Memory achieves a relatively balanced profile, especially on transfer- and retention-related dimensions, suggesting that organizing experiences into a more structured memory can help preserve and reuse information beyond raw instance retrieval. At the same time, its efficiency is constrained by graph construction, insight maintenance, and additional retrieval operations. DC-RS shows another form of trade-off. Its retrieve--update--solve pipeline can help the current task by reusing retrieved past history, but its fallback behavior may also weaken memory abstraction: when a new cheatsheet cannot be extracted, the retrieved history is used directly as the updated memory. This makes the memory useful as an ICL-style context for nearby tasks, but can also produce verbose final memories that are costly and less reliable for long-term generalization or retention.

Overall, this comparison reinforces the main message of \textsc{SeqMem-Eval}: memory quality is not determined by a single dominant score, but by how a memory mechanism balances adaptation, generalization, consolidation, retention, and cost. The profiles also suggest concrete design implications. Methods should move beyond raw instance storage toward mechanisms that abstract reusable experience, but such abstraction must be paired with validation and cost control; otherwise, stronger online performance may come primarily from additional inference or update budget rather than from more reliable memory evolution.

\section{Conclusion}

Sequentially evolving LLM memory is an important step toward agents that can accumulate experience and adapt beyond independent task solving. However, existing evaluations often reduce memory quality to a single aggregate score. Our study showed that this view is incomplete: similar aggregate performance can hide substantially different memory dynamics, like unstable online trajectories. 
We introduced \textsc{SeqMem-Eval}, a diagnostic evaluation framework that decomposes sequential memory behavior into online utility, hold-out generalization, backward transfer, forgetting, and efficiency. Across diverse tasks and memory methods, our analysis showed that these diagnostics provide a more complete view of memory behavior than aggregate scores alone. The resulting observations also offer useful insights for future memory research, highlighting the need to design and evaluate memory mechanisms in terms of generalization, transfer, retention, and efficiency.
Overall, \textsc{SeqMem-Eval} reframes LLM memory evaluation as the analysis of an evolving process rather than a final endpoint. We hope this perspective helps guide the development of more reliable, generalizable, and efficient memory-augmented LLM agents.



\bibliographystyle{abbrv}
\bibliography{refs}

\newpage
\appendix

\section{Detailed Experimental Setup}
\label{appx:exp-setup}

\subsection{Dataset configuration.}

We evaluate on a diverse set of benchmarks spanning programming, mathematical reasoning, factual and domain-specific reasoning, tool use, and embodied interaction. 
For code generation, we use \textbf{HumanEval}~\citep{chen2021evaluatinglargelanguagemodels}; for mathematical reasoning, we adopt \textbf{MATH500}~\citep{hendrycks2021measuringmathematicalproblemsolving}; for factual and domain-specific reasoning, we include \textbf{MMLU-Pro}~\citep{wang2024mmluprorobustchallengingmultitask}, focusing on advanced subjects such as engineering and physics; for tool-use and API grounding, we evaluate on \textbf{APIBench}~\citep{patil2023gorillalargelanguagemodel}; and for long-horizon goal-oriented interaction, we include \textbf{ALFWorld}~\citep{shridhar2021alfworldaligningtextembodied}. 
Together, these benchmarks cover both single-turn and multi-step settings, enabling evaluation of sequential memory across diverse forms of reasoning and interaction. 

For datasets without dedicated training splits (e.g., HumanEval and MMLU-Pro), we construct the ID hold-out set by taking the last 20\% of the test set. For APIBench, MATH500, and ALFWorld, the hold-out sets are sampled from the training data. The sampling is performed proportionally across task categories to preserve the original distribution, while ensuring that each category contains at least one task. For APIBench, we only use the HF subset, as the TF and TH subsets are relatively small and are nearly saturated by most models. Dataset statistics, including total size and hold-out split size, are summarized in Table~\ref{tab:dataset_stats}.

\begin{table}[ht]
\centering
\small
\caption{Dataset statistics, including total size and ID hold-out size.}
\label{tab:dataset_stats}
\begin{tabular}{l c c}
\toprule
\textbf{Dataset} & \textbf{Sequential test} & \textbf{Hold-out} \\
\midrule
HumanEval & 132 & 32 \\
MATH500 & 500 & 280 \\
MMLU-Pro-Engineering & 873 & 96 \\
MMLU-Pro-Physics & 1170 & 129 \\
APIBench-HF & 911 & 120 \\
ALFWorld & 140 & 24 \\
\bottomrule
\end{tabular}
\end{table}

\paragraph{Sequential Evaluation Protocol.}

All datasets are evaluated under a unified sequential protocol. Each example is processed once in sequence, and the memory state is incrementally updated after each interaction. To ensure fair comparison, all methods are evaluated using the same retrieval budget.  Each dataset is evaluated under a fixed task order shared by all methods. This design ensures that different memory mechanisms are compared under the same sequence of experiences. We emphasize, however, that task ordering is an intrinsic factor in sequential memory evaluation: different orders may expose different forms of adaptation, interference, and forgetting. Therefore, our reported results should be interpreted as diagnostics under controlled task streams rather than order-invariant estimates of memory quality. This protocol enables consistent measurement of online utility, hold-out generalization, backward transfer, forgetting, and efficiency across tasks and memory methods. 

\subsection{Model configuration.}
We conduct experiments using two representative large language models: Qwen3-8B~\citep{qwen3technicalreport} and MiniMax-M2.7~\citep{minimax2026m27}.

\paragraph{Qwen3-8B model configuration.}
We use the official recommended temperature setting of $0.7$ to balance generation quality and diversity. For HumanEval, MATH500, APIBench, and MMLU-Pro, reasoning is disabled, and the maximum number of generated tokens is set to $2048$ for all methods. For ALFWorld, due to its higher task complexity, we enable reasoning and increase the maximum number of generated tokens to $8192$ to allow for extended reasoning steps. All experiments are conducted on a single A100-80GB GPU.

\paragraph{MiniMax-M2.7 model configuration.}
We access the MiniMax model via the OpenRouter API. For all tasks and methods, we use a unified configuration with temperature set to $1.0$, reasoning level set to low, and the maximum number of generated tokens set to $16384$.

\subsection{Methods}
We evaluate a representative set of sequentially evolving LLM memory methods to systematically analyze memory behaviors under a unified sequential setting. 

(1) \textbf{Memory-free baseline.}
The memory-free baseline solves each task using only the underlying LLM without persistent memory. It serves as a reference point for measuring whether external memory provides benefits beyond the base model.

(2) \textbf{Instance-level experience retrieval.}
We include ExpRecent$_{k=10}$, ExpRecent$_{k=3}$, and ExpRAG$_{k=3}$ as lightweight baselines for raw experience reuse~\citep{wei2025evo}. 
ExpRecent$_k$ conditions on the most recent $k$ tasks, while ExpRAG$_k$ retrieves the top-$k$ most similar past tasks based on embedding similarity to the current input.
This comparison separates recency-based memory from relevance-based retrieval and evaluates whether instance-level experiences alone are sufficient for sequential adaptation without additional memory abstraction or refinement.

(3) \textbf{Structured and evolving memory.}
We evaluate Dynamic Cheatsheet Retrieval-and-Synthesis variant(DC-RS)~\citep{suzgun2025dynamiccheatsheettesttimelearning}, Agent Workflow Memory (AWM)~\citep{wang2024agentworkflowmemory}, and G-Memory~\citep{zhang2025gmemorytracinghierarchicalmemory}. 
These methods go beyond raw experience retrieval by organizing memory into higher-level structures such as cheatsheets, workflows, or hierarchical memory graphs. 
They allow us to examine whether structured memory representations improve transfer, retention, and generalization during sequential inference.

(4) \textbf{Experiential reflection.}
We include ExpeL~\citep{zhao2024expelllmagentsexperiential}, which derives reusable insights from successful and failed trajectories through reflection. Since the original formulation is not directly designed for our sequential evaluation protocol, we refactor it into a sequential memory setting. We evaluate its multi-try version as ExpeL-MT, following the original strategy of allowing multiple attempts per task, and additionally implement a single-try variant, ExpeL-ST, for controlled comparison. This comparison helps distinguish gains from experiential memory from those introduced by additional inference-time attempts.
Detailed algorithmic descriptions and implementation details for all methods are provided in Appendix~\ref{appx:baseline-alg}.


\paragraph{Method configuration.}
Method configurations are summarized in Table~\ref{tab:method_config}. For all methods involving retrieval, we standardize the pipeline by using Qwen3-0.6B-embedding~\citep{qwen3technicalreport} as the embedding model. Dense retrieval is performed by computing cosine similarity using the Sentence Transformers library~\cite{reimers2019sentence}.

\begin{table}[ht]
\centering
\small
\caption{Method-specific configurations used in sequential experiments.}
\label{tab:method_config}
\begin{tabular}{l l}
\toprule
\textbf{Method} & \textbf{Configuration} \\
\midrule
ExpRecent & \texttt{top-k=3},\texttt{top-k=10} \\
ExpRAG & \texttt{top-k=3} \\
AWM & \texttt{top-k=3}, \texttt{induce-steps=1} \\
DC-RS & \texttt{top-k=3} \\
ExpeL-MT & \texttt{top-k=3}, \texttt{max-tries=3}, \texttt{batch-update-size=8}, \texttt{max-num-rules=20} \\
ExpeL-ST & \texttt{top-k=3}, \texttt{batch-update-size=8}, \texttt{max-num-rules=20} \\
G-Memory & \texttt{successful-topk=2}, \texttt{failed-topk=1}, \texttt{insights-topk=10} \\
\bottomrule
\end{tabular}
\end{table}

\section{Practical Use of \textsc{SeqMem-Eval}}
\label{appx:diagnostic_usage}

\textsc{SeqMem-Eval} is not intended to collapse memory quality back into a single universal score. Different deployment settings may prioritize different aspects of memory behavior, such as online adaptation, hold-out generalization, retention, or computational cost. 
For example, an interactive agent may prioritize online utility and latency, whereas a long-term reasoning assistant may place more emphasis on hold-out generalization and forgetting.

As a practical guideline for method comparison, we suggest a two-stage procedure. First, identify methods that are Pareto competitive with respect to primary deployment constraints, such as \(\mathrm{OnlineAcc}\), \(\mathrm{HoldOutAcc}\), and efficiency. Second, among these candidates, use the diagnostic metrics to inspect memory dynamics: prefer methods with positive \(\mathrm{Trend}_{\mathrm{HO}}\), non-negative BWT at short and medium horizons, low forgetting, and low PED. This procedure preserves the diagnostic nature of the framework while providing a concrete way to compare methods without reducing memory quality to a single score.

\section{Additional Experimental Results}

\subsection{ID hold-out test results}
\label{appx:id-holdout}



\begin{figure}[ht]
    \centering
    \begin{subfigure}{\textwidth}
        \centering
        \includegraphics[width=\textwidth]{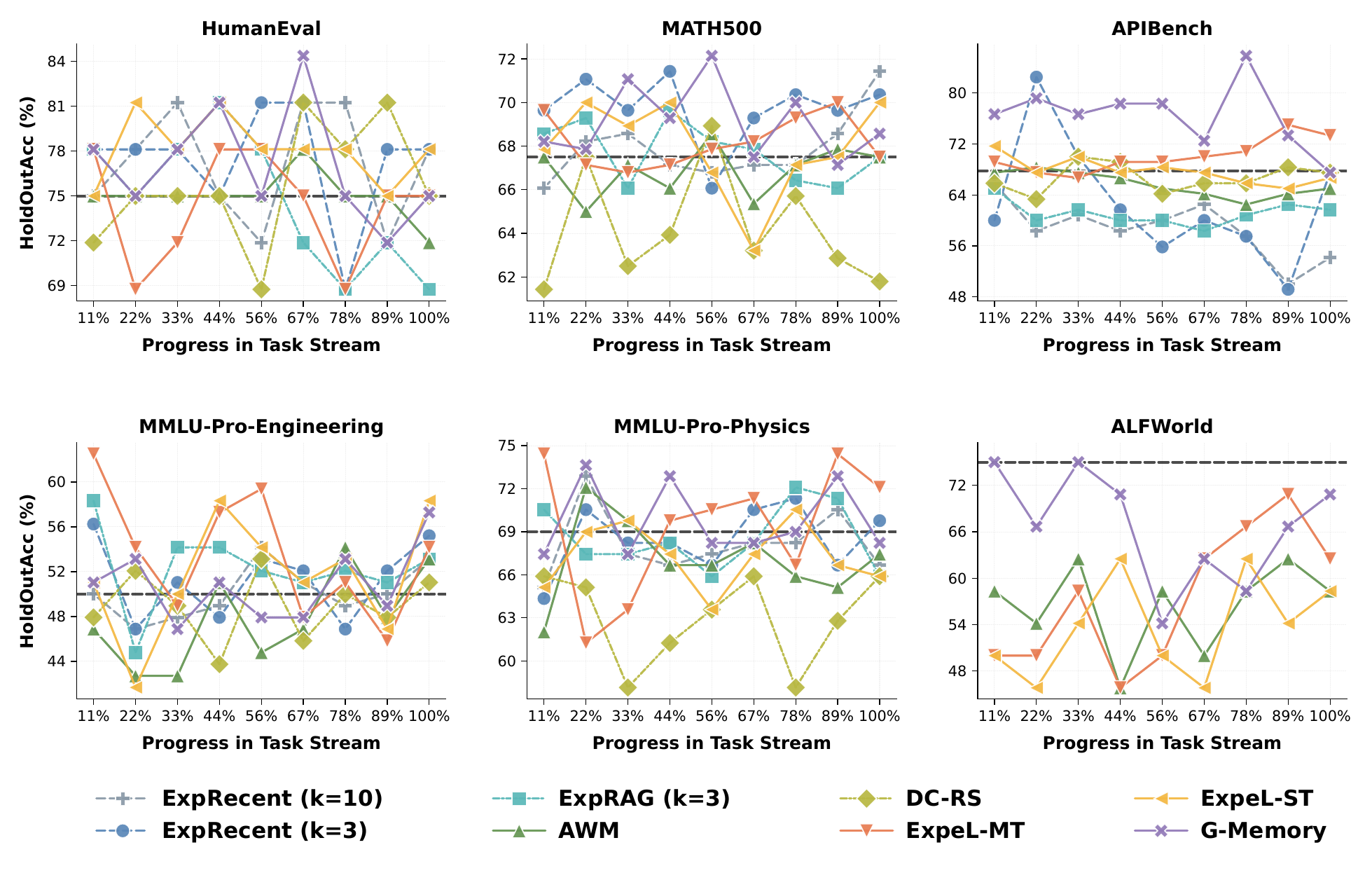}
        \caption{Qwen3-8B}
    \end{subfigure}

    \vspace{0.5em}

    \begin{subfigure}{\textwidth}
        \centering
        \includegraphics[width=\textwidth]{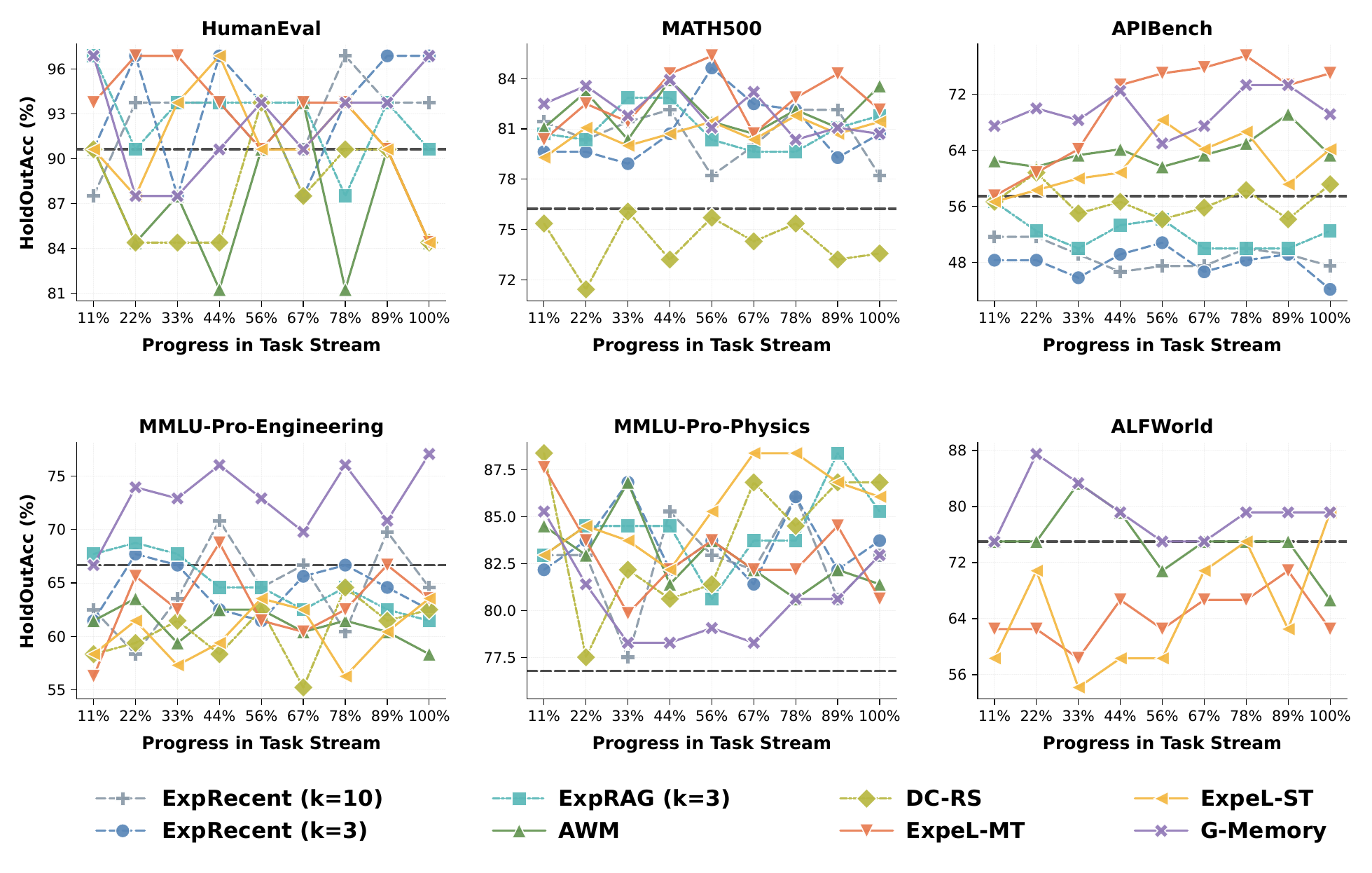}
        \caption{MiniMax-M2.7}
    \end{subfigure}

    \caption{In-distribution (ID) hold-out accuracy over sequential steps for different models.}
    \label{fig:holdout_id_models}
\end{figure}

\Cref{fig:holdout_id_models} present the full ID hold-out trajectories over sequential memory updates.
Consistent with the main-text analysis, hold-out generalization rarely improves monotonically as memory evolves. Across many datasets, trajectories fluctuate substantially around the memory-free baseline, and later memory states often fail to outperform earlier ones. This pattern is particularly pronounced on APIBench and MMLU-Pro-Engineering, where several methods exhibit transient improvements followed by degradation, suggesting that memory updates may overfit to recently observed interactions rather than accumulate transferable knowledge. In multiple cases, peak hold-out performance occurs at intermediate stages rather than in the final memory state, indicating that useful information can later be diluted, overwritten, or negatively affected by subsequent updates.

The trajectory-level analysis also reveals clear differences across memory mechanisms. Retrieval-based methods generally produce relatively stable, low-variance trajectories close to the baseline, implying limited long-term knowledge accumulation despite reduced instability. In contrast, stronger memory systems such as ExpeL-MT and G-Memory often achieve higher peak hold-out performance but exhibit highly non-stationary trajectories with substantial oscillations across memory steps. This trade-off suggests that aggressive memory synthesis and refinement may improve adaptability while reducing memory stability. Moreover, identical methods can exhibit qualitatively different hold-out dynamics across LLM backbones, indicating that memory generalization depends not only on stored experience but also on the backbone’s ability to consistently exploit evolving memory states. Overall, these results further support the central claim of RQ2: final hold-out accuracy alone obscures important temporal properties of memory generalization, including instability, reversibility, and sensitivity to later updates.

\subsection{OOD hold-out test results on Qwen3-8B}
\label{appx:od-holdout-qwen}

\begin{figure}[t]
    \centering
    \includegraphics[width=\textwidth]{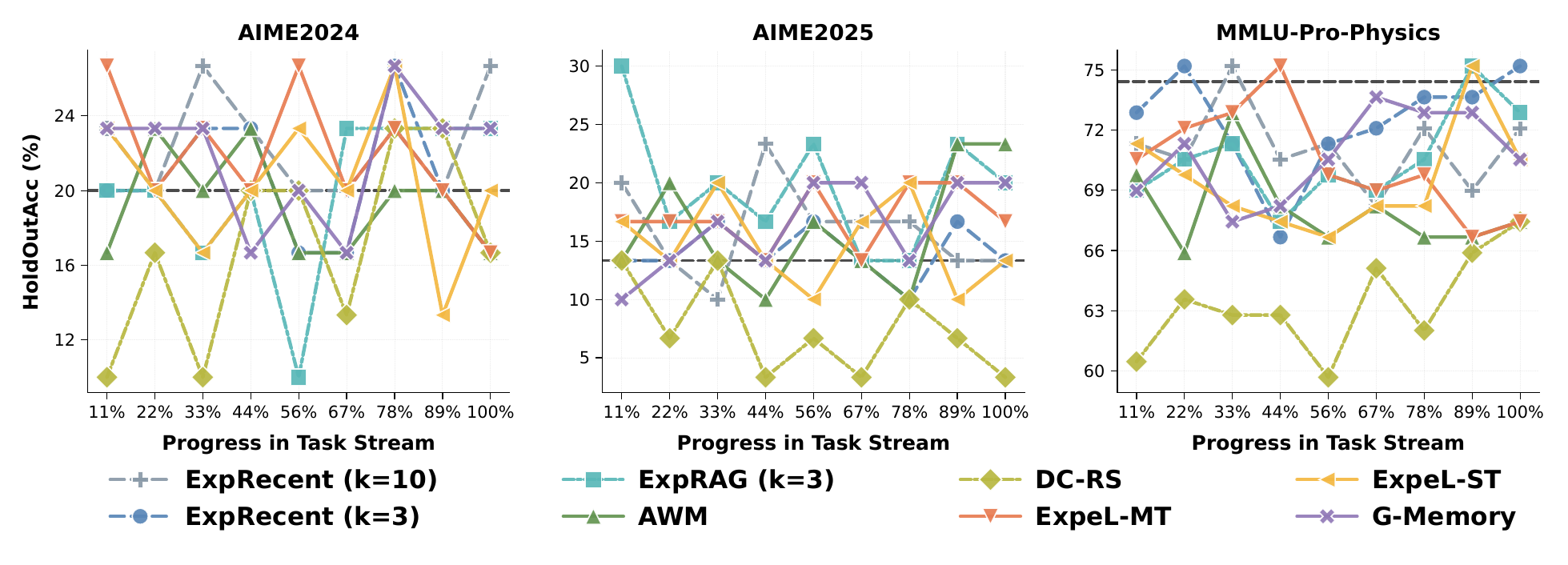}
    \caption{Out-of-distribution (OOD) hold-out accuracy over sequential steps for the Qwen3-8B model, evaluated on unseen tasks not encountered during the online updating process.}
    \label{fig:qwen_holdout_ood}
\end{figure}

We evaluate OOD generalization by transferring memory states learned from the MATH500 stream to AIME2024, AIME2025, and MMLU-Pro-Physics, with results shown in \Cref{fig:qwen_holdout_ood}. Compared with the ID setting, OOD trajectories exhibit weaker and less stable gains, indicating limited transferability of learned memory beyond the training distribution.

On AIME2024 and AIME2025, memory provides only marginal improvements over the memory-free baseline, likely due to partial overlap in mathematical reasoning patterns. However, these gains are inconsistent across methods and memory stages, suggesting that the learned memory captures limited transferable structure.
The degradation is more pronounced on MMLU-Pro-Physics, where performance remains consistently below the baseline, even for methods that construct higher-level synthesized insights. This behavior indicates weak cross-domain transfer and limited generalization of memory beyond the source domain.

Across all OOD datasets, hold-out trajectories exhibit substantial fluctuations, and later memory states frequently fail to outperform earlier ones. Although this instability mirrors the behavior observed under ID evaluation, it becomes more pronounced under distribution shift, highlighting the lack of stable and transferable memory evolution.

\subsection{Efficiency Results}
\label{appx:efficiency}

\begin{figure}[ht]
    \centering
    \begin{subfigure}{\textwidth}
        \centering
        \includegraphics[width=0.95\textwidth]{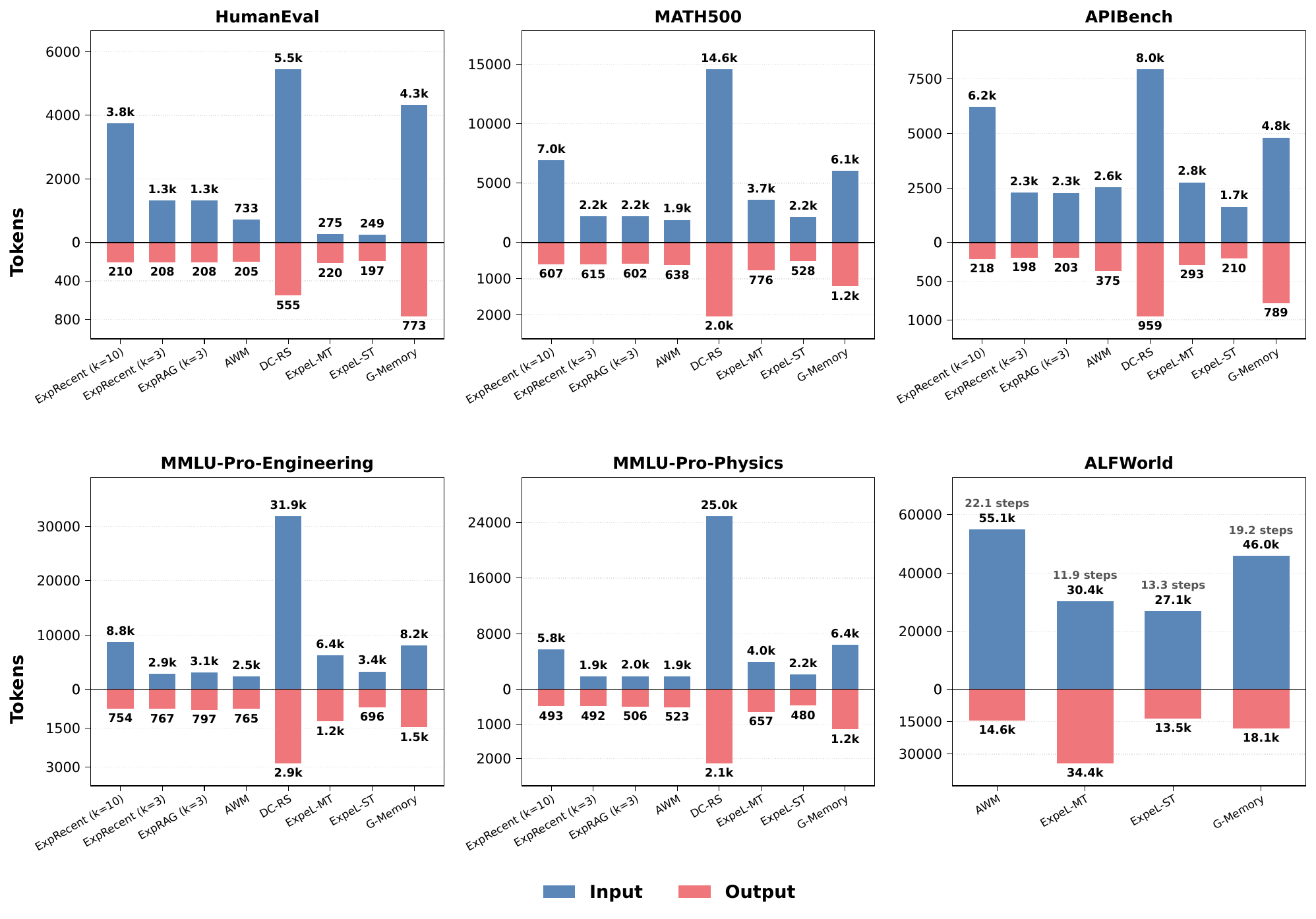}
        \caption{Token consumption.}
        \label{fig:qwen_token}
    \end{subfigure}

    \vspace{0.5em}

    \begin{subfigure}{\textwidth}
        \centering
        \includegraphics[width=0.95\textwidth]{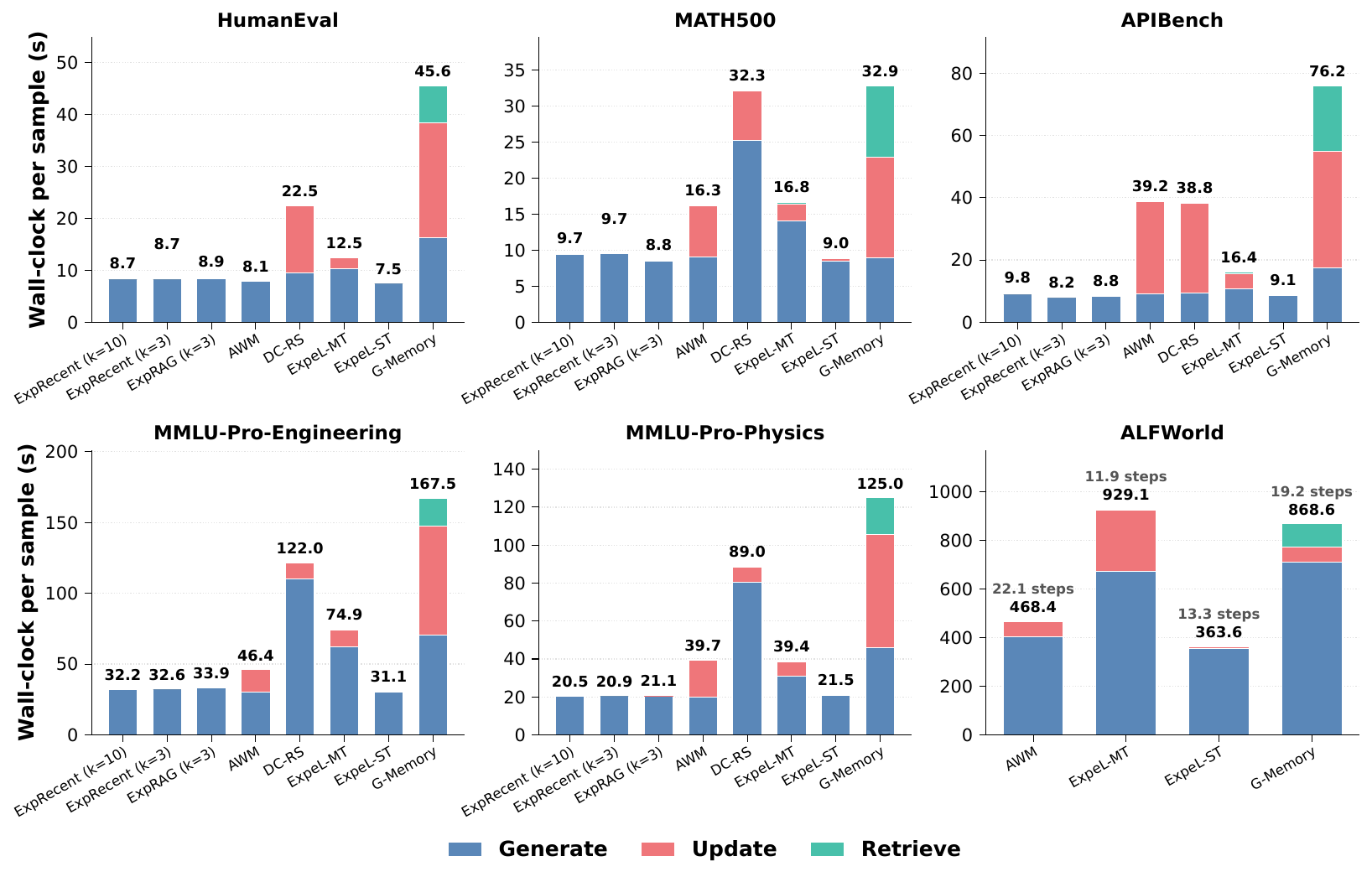}
        \caption{Runtime.}
        \label{fig:qwen_timing}
    \end{subfigure}

    \caption{Efficiency analysis of token consumption and runtime for the Qwen3-8B model.}
    \label{fig:qwen_efficiency}
\end{figure}

\begin{figure}[ht]
    \centering
    \begin{subfigure}{\textwidth}
        \centering
        \includegraphics[width=0.95\textwidth]{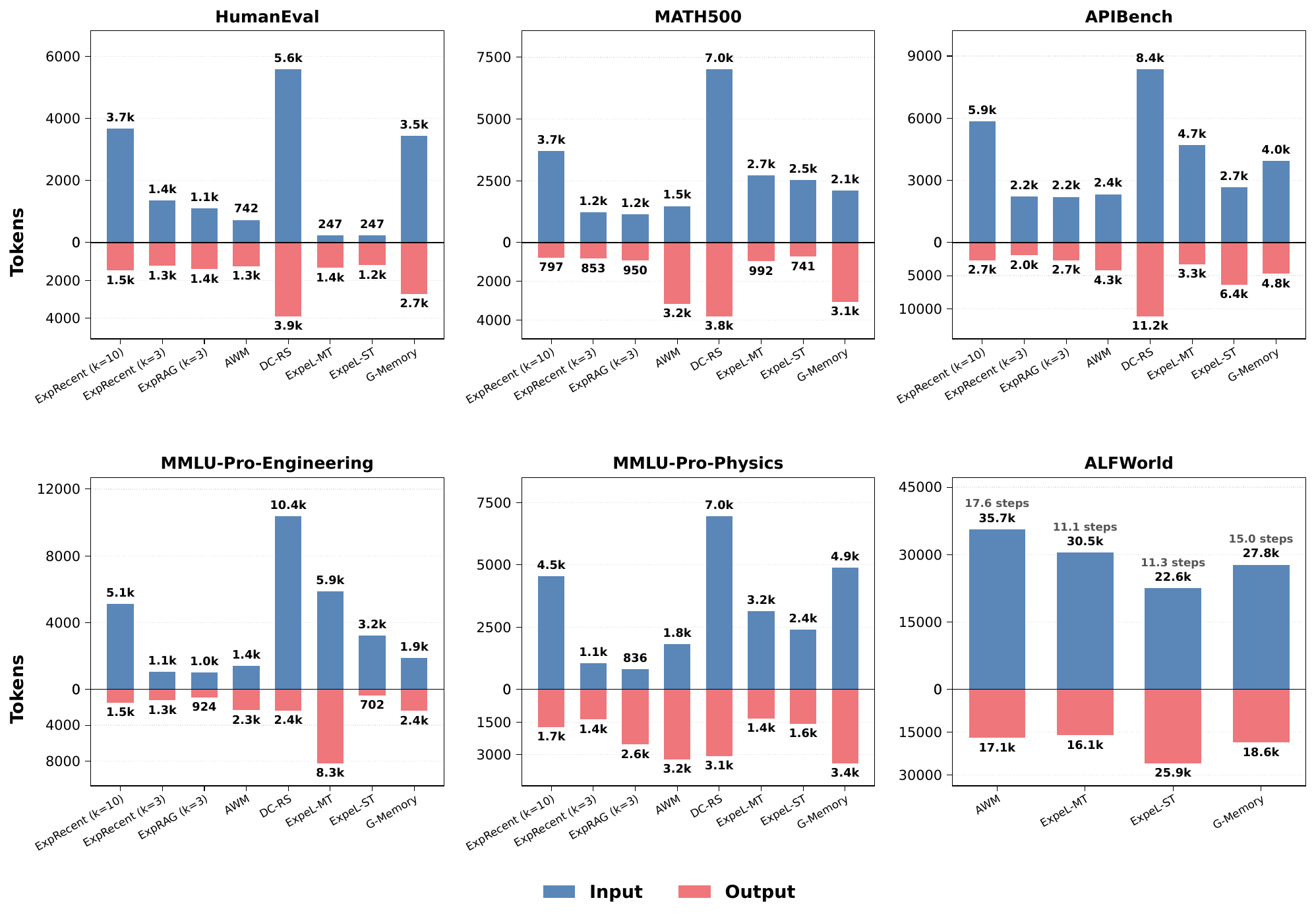}
        \caption{Token consumption.}
        \label{fig:minimax_token}
    \end{subfigure}

    \vspace{0.5em}

    \begin{subfigure}{\textwidth}
        \centering
        \includegraphics[width=0.95\textwidth]{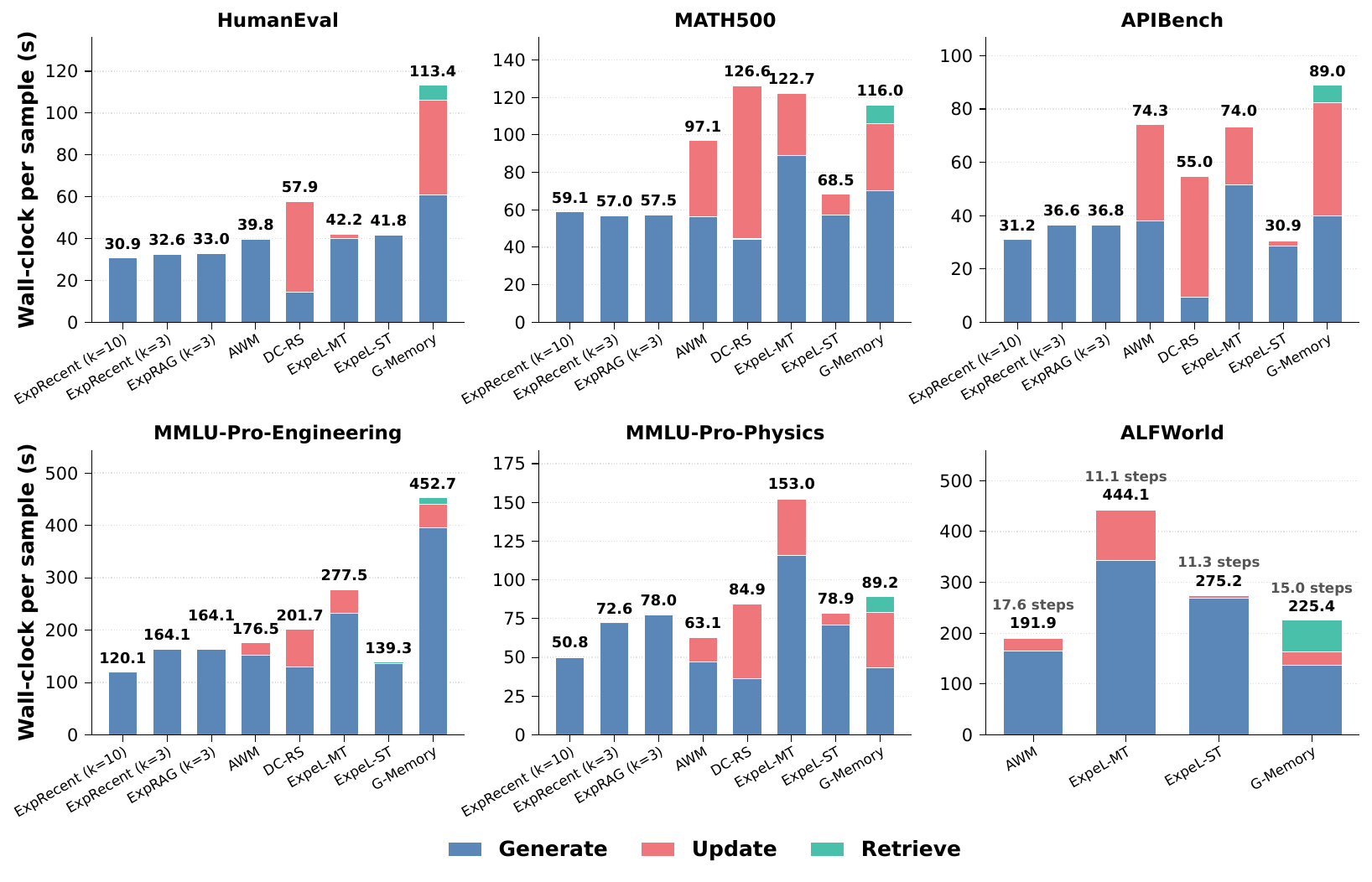}
        \caption{Runtime.}
        \label{fig:minimax_timing}
    \end{subfigure}

    \caption{Efficiency analysis of token consumption and runtime for the MiniMax-M2.7 model.}
    \label{fig:minimax_efficiency}
\end{figure}

Figures~\ref{fig:qwen_efficiency} and~\ref{fig:minimax_efficiency} present detailed runtime and token-consumption breakdowns across datasets and memory mechanisms for Qwen3-8B and MiniMax-M2.7. Consistent with the main-text analysis, runtime is dominated by generation, while token overhead primarily originates from memory construction and update operations. The breakdown further shows that different memory mechanisms induce distinct efficiency profiles. Retrieval-based methods such as ExpRecent and ExpRAG mainly increase input tokens through longer retrieved contexts while incurring relatively small update overhead, resulting in moderate and stable computational growth. In contrast, methods based on synthesized or structured memory introduce substantially larger update costs. DC-RS exhibits particularly high token consumption because each update requires long structured prompts for cheatsheet synthesis, causing update tokens to dominate total usage on several datasets. G-Memory shows a different overhead pattern: although its token growth is smaller than that of DC-RS, runtime increases substantially due to graph construction, insight maintenance, and additional retrieval operations during inference.

The dataset-level breakdown further indicates that efficiency depends strongly on task characteristics. On reasoning-intensive datasets such as MMLU-Pro-Engineering and MMLU-Pro-Physics, runtime increases disproportionately even under moderate token growth, suggesting that latency is affected not only by context length but also by more complex inference trajectories. ALFWorld exhibits the most distinct behavior: methods with iterative interaction or retrieval mechanisms incur extremely large runtime despite comparatively moderate token usage, indicating that multi-step environment interaction amplifies latency beyond pure language generation cost. In addition, identical memory mechanisms can exhibit substantially different efficiency profiles across LLM backbones. MiniMax-M2.7 generally produces higher runtime despite similar or lower token counts, implying that memory overhead is jointly determined by both the external memory algorithm and the backbone-specific inference characteristics. Overall, these results show that aggregate token usage alone is insufficient to characterize computational efficiency, as memory update complexity, retrieval structure, and interaction dynamics contribute to end-to-end cost differently.

\clearpage

\clearpage

\section{Algorithms}
\label{appx:baseline-alg}
\begin{algorithm}[ht]
\small
\caption{Dynamic Cheatsheet Retrieval-and-Synthesis (DC-RS)}\label{alg:DC} %
\begin{algorithmic}[1]
\REQUIRE Sequence of tasks $\gT=\{T_i\}_{i=1}^{N}$, large language model $\gL$, retrival budget $K$, generator template $P_{gen}$, curator template $P_{cur}$ 
\STATE \textbf{Initialize:} Cheatsheet $\gM_0 \gets \emptyset$, history $\gH_0 \gets \emptyset$
\FOR{Task $T_i=(x_i, y_i) \in \gT$}
    \IF{$|\gH_{i-1}|>0$}
            \STATE Retrieve Top-$K$ similar past cases $C_{retr}$ from $\gH_{i-1}$ \algcomment{Search}
        \ELSE
            \STATE $\mathcal{P} \gets \emptyset$
        \ENDIF
\STATE Update cheatsheet $\gM_{i} \gets \gL(P_{cur},\; C_{retr},\; x_i,\; \gM_{i-1})$ \algcomment{Evolve}
\STATE Answer $\hat{y}_{i} \gets \gL(P_{gen},\; x_i,\; \gM_{i})$ \algcomment{Synthesis}
\STATE Update history $\gH_{i} \gets \gH_{i-1}\cup \{x_i,\hat{y}_{i}\}$ \algcomment{Evolve}
\ENDFOR
\end{algorithmic}
\end{algorithm}

\begin{algorithm}[ht]
\caption{Agent Workflow Memory (AWM)}
\label{alg:awm-online-induce}
\begin{algorithmic}[1]
\REQUIRE Sequence of tasks $\gT=\{T_i\}_{i=1}^{N}$, large language model $\gL$, induction prompt $P_{in}$, one-shot prompt $P_{one}$
\STATE \textbf{Initialize:} base memory $\gM_0 \gets \emptyset$\

\FOR{Task $T_i=(x_i, y_i) \in \gT$}
    \STATE Answer $\hat{y}_{i} \gets \gL(x_i, \gM_{i-1})$ and get experience $e_i \gets (x_i, \hat{y}_{i})$ \algcomment{Synthesis}
    \STATE $success \gets \mathrm{Evaluator}(e_i)$

    \IF{$success = 1$}
        \STATE $w_i \gets \gL(P_{in} \;\Vert\; P_{one} \;\Vert\; e_i \;\Vert\; \texttt{"\#\# Summary Workflows"})$ \algcomment{Evolve}
        \STATE $\gM_{i} \gets \gM_{i-1} \cup w_i$ \algcomment{Evolve}
    \ELSE
        \STATE $\gM_{i} \gets \gM_{i-1}$
    \ENDIF
\ENDFOR

\end{algorithmic}
\end{algorithm}

\begin{algorithm}[ht]
\caption{ExpeL-ST}
\label{alg:expel-online-onetry}
\begin{algorithmic}[1]
\REQUIRE Sequence of tasks $\gT=\{T_i\}_{i=1}^{N}$, large language model $\gL$, insight extraction prompt $P_{in}$, retrival budget $K$, step size $L$
\STATE \textbf{Initialize:} Experience pool $\gB \gets F_{\textsc{manual}}$ (seed few-shot demos if available), recent success trajectories $\gS \gets \emptyset$, insight set $\hat{\iota} \gets \emptyset$

\FOR{Task $T_i=(x_i, y_i) \in \gT$}
    \STATE Retrieve Top-$K$ similar success cases $F_{\textsc{sim}}$ from $\gB$ \algcomment{Search}
    \STATE Answer $\hat{y}_{i} \gets \gL(x_i, F_{\textsc{sim}}, \hat{\iota})$ \algcomment{Synthesis}

    \STATE $success \gets \mathrm{Evaluator}(x_i, y_i, \hat{y}_{i})$

    \IF{$success = 1$}
        \STATE $\gS \gets \gS \cup \{(x_i, \hat{y}_{i})\}$  
        \STATE $\gB \gets \gB \cup \{(x_i, \hat{y}_{i})\}$ \algcomment{Evolve}
        \IF{$|\gS| = L$} 
        \STATE $\hat{\iota} \leftarrow \gL(P_{in}, \gS, \hat{\iota})$ \algcomment{Evolve}
        \STATE $\gS \gets \emptyset$
        \ENDIF
    \ENDIF
\ENDFOR
\end{algorithmic}
\end{algorithm}

\begin{algorithm}[ht]
\caption{ExpeL-MT}
\label{alg:expel-online-multitry}
\begin{algorithmic}[1]
\REQUIRE Sequence of tasks $\gT=\{T_i\}_{i=1}^{N}$, large language model $\gL$,
self-reflection model $\gL_{\textsc{reflect}}$,
insight prompt $P_{in}$, retrieval budget $K$, step size $L$,
max tries $Z$
\STATE \textbf{Initialize:} Experience pool $\gB \gets F_{\textsc{manual}}$ (seed demos if available),
recent success set $\gS \gets \emptyset$, insight set $\hat{\iota} \gets \emptyset$

\FOR{Task $T_i=(x_i, y_i) \in \gT$}
    \STATE Retrieve Top-$K$ similar success cases $F_{\textsc{sim}}$ from $\gB$ \algcomment{Search}
    \STATE $\nu \gets \texttt{""}$ 
    \STATE $\tau^{\textsc{succ}} \gets \emptyset$, $\tau^{\textsc{fail}} \gets \emptyset$

    \FOR{$z=1$ \TO $Z$}
        \STATE $\hat{y}_i^{(z)} \gets \gL(x_i, F_{\textsc{sim}}, \hat{\iota}, \nu)$ \algcomment{Synthesis}
        \STATE $success^{(z)} \gets \mathrm{Evaluator}(x_i, y_i, \hat{y}_i^{(z)})$

        \IF{$success^{(z)} = 1$}
            \STATE $\tau^{\textsc{succ}} \gets (x_i, \hat{y}_i^{(z)})$
            \STATE $\gB \gets \gB \cup \{\tau^{\textsc{succ}}\}$ \algcomment{Evolve}
            \STATE \textbf{break}
        \ELSE
            \STATE $\tau^{\textsc{fail}} \gets (x_i, \hat{y}_i^{(z)})$ 
            \STATE $\nu \gets \textsc{Concat}(\nu,\; \gL_{\textsc{reflect}}(\tau^{\textsc{fail}}))$ \algcomment{Reflect}
        \ENDIF
    \ENDFOR

    \IF{$\tau^{\textsc{succ}} \neq \emptyset$}
        \STATE $\gS \gets \gS \cup \{\tau^{\textsc{succ}}\}$
    \ENDIF

    \IF{$\tau^{\textsc{succ}} \neq \emptyset \ \AND\  \tau^{\textsc{fail}} \neq \emptyset$}
        \STATE $\hat{\iota} \leftarrow \gL(P_{in}, \tau^{\textsc{succ}}, \tau^{\textsc{fail}}, \hat{\iota})$ \algcomment{Pair Update}
    \ENDIF

    \IF{$|\gS| = L$}
        \STATE $\hat{\iota} \leftarrow \gL(P_{in}, \gS, \hat{\iota})$ \algcomment{Batch Update}
        \STATE $\gS \gets \emptyset$
    \ENDIF
\ENDFOR
\end{algorithmic}
\end{algorithm}

\begin{algorithm}[ht]
\caption{G-Memory}
\label{alg:gmemory}
\begin{algorithmic}[1]
\REQUIRE Task stream $\gT$, MAS $\gG$,
query graph $\gG_{\textsc{qry}}=(\gQ,\gE_{\textsc{qry}})$
with query nodes $\gQ=\{q_i\}$,
insight graph $\gG_{\textsc{ins}}=(\gI,\gE_{\textsc{ins}})$,
budgets $K,M$

\FOR{$(Q_i,y_i)\in\gT$}

    \STATE $\gQ^S \gets
    \textsc{Retrieve}(Q_i,\gQ,K)$   
    \algcomment{Search}

    \STATE $\widetilde{\gQ}^S \gets
    \textsc{Expand}(\gQ^S,\gG_{\textsc{qry}})$
    \algcomment{Search}

    \STATE $\gI^S \gets
    \Pi_{Q\rightarrow I}(\widetilde{\gQ}^S,\gG_{\textsc{ins}})$
    \algcomment{Search}

    \STATE $\gQ^R \gets
    \textsc{Select}(\widetilde{\gQ}^S,M)$

    \STATE $\widehat{\gG}_{\textsc{inter}}
    \gets \textsc{Sparsify}(\gQ^R,Q_i)$

    \FOR{$C_j\in\gG$}
        \STATE $Mem_j \gets
        \Phi(\gI^S,
        \widehat{\gG}_{\textsc{inter}},
        Role_j,Q_i)$
    \ENDFOR

    \STATE $\hat y_i,\tau_i \gets \gG(Q_i)$
    \algcomment{Synthesis}

    \STATE $\Psi_i \gets
    \mathrm{Evaluator}(\hat y_i,y_i)$

    \STATE $\gG^{(Q_i)}_{\textsc{inter}}
    \gets \textsc{Trace}(\tau_i)$

    \STATE $\gG_{\textsc{qry}}
    \gets
    \textsc{UpdateQueryGraph}
    (\gG_{\textsc{qry}},Q_i,\Psi_i,\gG^{(Q_i)}_{\textsc{inter}},\gQ^R,\gI^S)$ \algcomment{Evolve}

    \STATE $\gG_{\textsc{ins}}
    \gets
    \textsc{UpdateInsightGraph}
    (\gG_{\textsc{ins}},\gI^S,\gG^{(Q_i)}_{\textsc{inter}},\Psi_i)$ \algcomment{Evolve}

\ENDFOR
\end{algorithmic}
\end{algorithm}

\clearpage

\section{Prompts}
\label{appx:prompt}

\newtcolorbox{promptbox}[1][]{
  breakable,
  colback=lightgraybox,
  colframe=black!40,
  title=\textbf{Evaluation Rubric Prompt},
  fonttitle=\bfseries,
  fontupper=\ttfamily\footnotesize,
  left=1mm,right=1mm,top=1mm,bottom=1mm,
  #1
}

\begin{promptbox}[title=\textbf{AIME2024 / AIME2025 Prompt}]
Solve the following AIME problem. The final answer is a non-negative integer between 0 and 999.

Think step by step. Keep the reasoning concise but complete.
At the very end, output ONLY the final integer inside a LaTeX box, e.g., \texttt{\textbackslash boxed\{42\}}.
Do not wrap anything except the integer answer inside \texttt{\textbackslash boxed\{\}}.

Problem: \{question\}
\end{promptbox}

\begin{promptbox}[title=\textbf{MATH500 Prompt}]
Solve the following math problem.

Please do your derivation in LaTeX as much as possible. Keep the reasoning concise but complete.
At the end, put only the final answer in the LAST \texttt{\textbackslash boxed\{...\}}.
If the final answer is a fraction, use LaTeX fraction form like \texttt{\textbackslash frac\{a\}\{b\}} (not decimal approximation unless required).

Problem: \{question\}
\end{promptbox}

\begin{promptbox}[title=\textbf{MMLU-Pro Prompt}]
Answer the following multiple-choice \{subject\} question. Exactly one option is correct.

Think step by step. Keep the reasoning concise but complete.
At the very end, output ONLY the letter of the correct option inside a LaTeX box, e.g., \texttt{\textbackslash boxed\{C\}}.
Do not wrap anything except the single letter A--J inside \texttt{\textbackslash boxed\{\}}.

Question: \{question\}

Options:
\{options\}
\end{promptbox}

\begin{promptbox}[title=\textbf{HumanEval Prompt}]
Complete the following Python function. Keep the given signature and docstring unchanged; only fill in the body (you may add helper functions above it if needed). Your function must be named exactly \texttt{\{entry\_point\}}.

\texttt{```python}

\{question\}

\texttt{```}

You may reason briefly first, but only the code will be executed against hidden unit tests. Put your FINAL complete function inside a single fenced block:

\texttt{```python}

\# include the full \texttt{def \{entry\_point\}(...)} here, plus any helpers it depends on

\texttt{```}

Do not output any other code fence. No commentary after the code block.

ANSWER:
\end{promptbox}

\begin{promptbox}[title=\textbf{ALFWorld Prompt}]
Imagine you are an intelligent agent in a household environment, and your goal is to perform actions to complete the task. At the beginning of the interaction, you will be given a detailed description of the environment and the goal.

For each turn, think briefly and then output your next action.

The available actions are:
\begin{itemize}
    \item go to \{recep\}
    \item take \{obj\} from \{recep\}
    \item move \{obj\} to \{recep\}
    \item open \{recep\}
    \item close \{recep\}
    \item use \{obj\}
    \item clean \{obj\} with \{recep\}
    \item heat \{obj\} with \{recep\}
    \item cool \{obj\} with \{recep\}
\end{itemize}
where \{obj\} and \{recep\} correspond to objects and receptacles.

After each turn, the environment will provide feedback. If the feedback is ``Nothing happened'', the previous action is invalid and you should try a different action.

Your response must follow this format:
\begin{itemize}
    \item Thought:
    \item Action:
\end{itemize}
\end{promptbox}

\begin{promptbox}[title=\textbf{APIBench Prompt}]
You are a helpful API writer who can write APIs based on requirements.

\{question\}

Write a Python program in 1 to 2 lines to call API in \{framework\}.

The answer should follow the format: 
\verb|<<<|domain\verb|>>>| \$DOMAIN, \verb|<<<|api\_call\verb|>>>|: \$API\_CALL, \verb|<<<|api\_provider\verb|>>>|: \$API\_PROVIDER, \verb|<<<|explanation\verb|>>>|: \$EXPLANATION, \verb|<<<|code\verb|>>>|: \$CODE.
Here are the requirements:

\begin{enumerate}
    \item \texttt{\$DOMAIN} should be inferred from the task description.
    \item \texttt{\$API\_CALL} should have only one line of code that calls the API.
    \item \texttt{\$API\_PROVIDER} should be the programming framework used.
    \item \texttt{\$EXPLANATION} should be a step-by-step explanation.
    \item \texttt{\$CODE} is the Python code.
    \item Do not repeat the format in your answer.
\end{enumerate}
\end{promptbox}

\section{Case Study: Useful Memory State Destroyed by Subsequent Updates}
\label{appx:case-study}

This case study isolates a representative failure mode in sequentially evolving memory systems: a memory state that is initially useful gradually becomes corrupted or overwritten by later updates, eventually reducing downstream performance despite continued exposure to in-distribution data. We analyze ExpeL-MT on the MMLU-Pro-Engineering benchmark using Qwen3-8B as the backbone model, and focus on a single hold-out question on which the model initially answers correctly and progressively fails as the memory evolves. The full reconstruction is based on intermediate memory snapshots, retrieved trajectories, and checkpoint-level evaluation logs.

Hold-out evaluation is conducted with \texttt{max-tries=1} so that each checkpoint reflects the quality of the memory state itself rather than additional retry-and-reflect behavior.

\subsection{The Hold-out Question}

\begin{promptbox}
[title=\textbf{Question 87:}]
A 1$\frac{1}{2}$-in schedule-40 steam pipe is laid in the atmosphere where the temperature is 50~$^\circ$F. The steam inside it is saturated at 100 psia. Consider the pipe to be a grey body and uninsulated. The coefficient of heat transfer by natural convection from the outside surface is 2.0 Btu/(hr$\cdot$ft$^2\cdot^\circ$R). Calculate the amount of steam condensed per hour per unit length of pipe.

\vspace{0.3em}

A. 0.65 \quad
B. 0.50 \quad
\textbf{C. 0.70} \quad
D. 0.55 \quad
E. 0.40

F. 0.80 \quad
G. 0.75 \quad
H. 0.90 \quad
I. 1.00 \quad
\textbf{J. 0.60}
\end{promptbox}

The question belongs to the \textbf{Heat Transfer / Thermodynamics} category. 
The standard solution accounts for both natural convection and thermal radiation from the pipe surface. 
At 100 psia, saturated steam has $T_s \approx 327.8^\circ$F and $h_{fg}\approx 880~\mathrm{Btu/lbm}$. 
Using the 1.5-in schedule-40 outside diameter $D_o\approx1.90$ in $=0.158~\mathrm{ft}$, the outside area per foot is
\[
A=\pi D_o L \approx \pi \cdot 0.158 \cdot 1 = 0.497~\mathrm{ft^2/ft}.
\]
The convective heat loss is
\[
\dot Q_{\mathrm{conv}}
=
hA(T_s-T_\infty)
=
2.0\cdot 0.497 \cdot (327.8-50)
\approx 276~\mathrm{Btu/(hr\cdot ft)}.
\]
Because the pipe is grey-body and uninsulated, radiation must also be included:
\[
\dot Q_{\mathrm{rad}}
=
\epsilon\sigma A(T_s^4-T_\infty^4).
\]
Using a typical grey-body emissivity for an oxidized metal surface, the radiative loss is on the same order as convection, yielding total heat loss about
\[
\dot Q_{\mathrm{tot}}\approx 530~\mathrm{Btu/(hr\cdot ft)}.
\]
Therefore,
\[
\dot m
=
\frac{\dot Q_{\mathrm{tot}}}{h_{fg}}
\approx
\frac{530}{880}
\approx
0.60~\mathrm{lb/(hr\cdot ft)}.
\]
Thus, the standard answer is $\boxed{\text{J}}$.

\subsection{Memory Evolution and Prediction Transition}

We reconstruct the full 20-slot ExpeL-MT insight pool from intermediate memory snapshots to analyze how thermo-related rules evolve throughout the sequential stream. A rule is considered thermo-related if it contains terms such as \textit{steam, condensation, enthalpy, heat transfer, latent heat,} or $h_{fg}$. Representative early-stage rules include: 

\begin{itemize}[leftmargin=1.2em]
\item
\textit{``Calculate heat transferred using final and initial enthalpy, considering temperature changes, specific heat capacities, and molecular weight for unit conversion.''}

\item
\textit{``Calculate exit temperature of a fluid in a heat exchanger...''}

\item
\textit{``Calculate initial quality...''}
\end{itemize}

During the early phase of the stream, which primarily contains Thermodynamics and Heat Transfer questions, the bounded insight pool gradually accumulates thermo-related rules. However, after the stream transitions to electrical engineering and communication topics, newly inserted rules progressively overwrite the earlier thermo-related insights.

ExpeL-MT relies on two memory sources during inference: (1) retrieved top-$k$ trajectories from the experience pool, and (2) an evolving insight pool produced through reflection and summarization. To identify the source of failure, we reconstruct both components across intermediate memory stages. Table~\ref{tab:prediction_trajectory} summarizes the prediction trajectory together with the number of thermo-related rules retained in memory. Prediction correctness strongly correlates with thermo-rule retention: the model remains mostly correct while thermo-related insights are preserved, but becomes consistently unstable once these rules disappear after $K=582$.

Importantly, retrieval quality does not degrade over time. Both early and late checkpoints retrieve highly relevant steam-condensation trajectories, and the retrieval similarity score increases from
\[
0.678 \rightarrow 0.753.
\]
Thus, the retrieved top-$k$ trajectories remain semantically relevant throughout the stream, indicating that the downstream degradation cannot be attributed to retrieval failure.

Instead, the failure originates from corruption within the evolving insight pool. As thermo-related insights are progressively replaced by unrelated electrical-engineering rules, the model loses the thermo-specific contextual anchors required for stable reasoning. The transition approximately coincides with the disappearance of thermo-related rules from the bounded insight pool. By the final checkpoint, all 20 insight slots are occupied by unrelated rules involving SCR currents, signal-processing formulas, torque equations, capacitance relations, and Reynolds-number similarity constraints. These observations suggest that the degradation is primarily caused by destructive memory updates rather than insufficient retrieval quality or backbone reasoning capability.

\begin{table}[t]
\centering
\small
\caption{
Prediction trajectory of the hold-out thermodynamics question under evolving ExpeL-MT memory states.
}
\label{tab:prediction_trajectory}
\setlength{\tabcolsep}{4pt}
\renewcommand{\arraystretch}{1.1}

\begin{tabular}{c c c c p{4.6cm}}
\toprule
\textbf{Checkpoint $K$}
& \textbf{Pred.}
& \textbf{Correct?}
& \textbf{\# Thermo Rules}
& \textbf{Model Behavior} \\
\midrule

97
& J
& \cmark
& 5
& Uses correct thermo constants and computes $\dot m \approx 0.60$, mapped to J. \\
\midrule

194
& D
& \xmark
& 6
& Mis-applies LMTD and obtains an incorrect temperature difference. \\
\midrule

291
& J
& \cmark
& 4
& Same reasoning path as $K=97$. \\
\midrule

388
& J
& \cmark
& 3
& Correct thermo grounding preserved. \\
\midrule

485
& J
& \cmark
& 1
& Final checkpoint with stable thermo-context grounding. \\
\midrule

582
& A
& \xmark
& 0
& Thermo-related insights disappear; latent heat value drifts and $\dot m$ shifts downward. \\
\midrule

679
& E
& \xmark
& 0
& Saturation temperature also drifts after thermo-context removal. \\
\midrule

776
& C
& \xmark
& 0
& Partial recovery of thermo constants but incorrect option selection. \\
\midrule

873
& B
& \xmark
& 0
& Prediction remains unstable after complete thermo-context overwrite. \\

\bottomrule
\end{tabular}
\end{table}

\subsection{Failure Mechanism and Implication}

The single-item trajectory closely mirrors the aggregate hold-out behavior across all 96 Engineering questions: strong early improvement is followed by gradual erosion and eventual regression toward the no-memory baseline. The case study therefore reflects a broader structural failure mode rather than an isolated prediction error.

This degradation emerges from the interaction between bounded memory capacity and sequential subject-localized updates. In ExpeL-MT, the insight pool is capped at:
\[
\texttt{max-num-rules}=20.
\]
Once the pool becomes saturated, newly generated rules implicitly replace or edit earlier entries. However, the eviction process does not estimate downstream utility, subject importance, or future dependency. As a result, memory retention becomes dominated by recency rather than long-term usefulness.

The effect is amplified by the subject-blocked ordering of the online stream. Neighboring tasks frequently belong to the same subject category, causing memory updates to become highly concentrated within local domains. During the early phase of the stream, thermo-related rules accumulate and stabilize prediction behavior on the hold-out Heat Transfer question. However, after the stream transitions into electrical-engineering and communication topics, newly inserted rules progressively overwrite the earlier thermo-related insights.

Importantly, the failure is not caused by retrieval degradation. Retrieved top-$k$ trajectories remain highly relevant throughout the stream, and retrieval similarity even increases over time. Instead, the degradation originates from corruption within the bounded insight pool itself. Once thermo-related rules disappear from memory, the model loses the thermo-specific contextual anchors required for stable reasoning, leading to drifting constants, unstable intermediate calculations, and eventually persistent prediction failure.

Consequently, memory updates become competitive rather than accumulative: later subjects overwrite previously useful subject-specific context instead of consolidating it. Together, these dynamics produce a clear form of:

\begin{center}
\fcolorbox{black!40}{gray!10}{
\parbox{0.9\linewidth}{
\centering
\textbf{catastrophic forgetting through bounded memory eviction}
}
}
\end{center}

More broadly, this case suggests that long-horizon memory systems likely require utility-aware retention, adaptive eviction policies, subject-balanced allocation, or retrieval-time filtering over larger persistent stores. Otherwise, useful contextual anchors accumulated early in the stream may be systematically overwritten by later updates, leading to transient gains followed by irreversible degradation.

\section{Limitations}
\label{appx:limitation}
\paragraph{Task-order sensitivity.}
A limitation of our current study is that most of the main experiments are conducted under a fixed task order for each dataset. This choice enables controlled comparison across memory methods, but it does not fully characterize uncertainty induced by different sequential orders. In principle, \textsc{SeqMem-Eval} can be applied to multiple randomized task streams and report confidence intervals for the metrics. In practice, this is computationally expensive because each task order requires rerunning the full sequential memory process, including retrieval, memory updates, hold-out evaluation at intermediate checkpoints, and retrospective evaluation for BWT and forgetting. For LLM-based memory methods such as reflection- or synthesis-based approaches, this cost is further amplified by additional API calls and generated tokens. We therefore leave large-scale multi-order uncertainty estimation as an important direction for future benchmark extensions.

\paragraph{Scope of memory mechanisms.}
Our evaluation focuses primarily on prompt-based sequentially evolving memory, where the base LLM remains fixed and memory is updated through retrieval, summarization, reflection, workflows, or other external text-based mechanisms. This scope matches many current memory-augmented agent systems and allows us to compare diverse methods under a unified test-time protocol. However, another emerging direction is training-based evolution, where experience is used not only to update an external memory state but also to improve policies, skills, or model behavior through learning. For example, recent work such as SkillRL~\cite{xia2026skillrl} studies recursive skill-augmented reinforcement learning, where a skill library co-evolves with the agent's policy during RL optimization. Such training-based approaches may exhibit different memory dynamics, including longer-term policy adaptation, different forms of transfer, and different efficiency trade-offs. While \textsc{SeqMem-Eval}'s diagnostic perspective may still be useful for analyzing these systems, our current experiments do not directly evaluate training-based evolving agents. Extending the framework to jointly assess external memory evolution and parameter- or policy-level adaptation is an important direction for future work.

\paragraph{Evaluation cost of diagnostic metrics.}
Although \textsc{SeqMem-Eval} provides a more fine-grained view of memory behavior, some of its metrics require additional evaluation beyond standard online accuracy. For example, hold-out generalization requires evaluating intermediate memory states on unseen tasks, while BWT and forgetting require revisiting previously encountered tasks under later memory states. These retrospective evaluations increase the total number of LLM calls, especially for long task streams or methods with expensive memory updates. In this work, we mitigate this cost by evaluating BWT and forgetting over selected temporal horizons and by sampling memory checkpoints for hold-out evaluation. Future work could explore more efficient approximations, such as adaptive checkpoint selection, smaller diagnostic probes, or surrogate estimators for memory quality.

\section{Broader Impacts}\label{appx:broader}

In this paper, we propose \textsc{SeqMem-Eval}, a diagnostic evaluation framework for sequential LLM memory. Our work aims to provide a more comprehensive understanding of how memory-augmented LLMs behave over time. We believe better evaluation of memory behavior may help identify failure modes before deployment and support safer use of LLM agents in long-term interactive settings. While we emphasize the importance of responsible use, we do not anticipate any major negative societal impacts from our work.





\end{document}

%% file: math_commands.tex
\usepackage{amsmath,amsfonts,bm,amsthm}

\def\eqref#1{equation~\ref{#1}}

\def\1{\bm{1}}

\DeclareMathAlphabet{\mathsfit}{\encodingdefault}{\sfdefault}{m}{sl}
\SetMathAlphabet{\mathsfit}{bold}{\encodingdefault}{\sfdefault}{bx}{n}

\def\gB{{\mathcal{B}}}

\def\gE{{\mathcal{E}}}

\def\gG{{\mathcal{G}}}
\def\gH{{\mathcal{H}}}
\def\gI{{\mathcal{I}}}

\def\gL{{\mathcal{L}}}
\def\gM{{\mathcal{M}}}

\def\gQ{{\mathcal{Q}}}

\def\gS{{\mathcal{S}}}
\def\gT{{\mathcal{T}}}



%% file: refs.bib
@misc{chen2021evaluatinglargelanguagemodels,
      title={Evaluating Large Language Models Trained on Code}, 
      author={Mark Chen and Jerry Tworek and Heewoo Jun and Qiming Yuan and Henrique Ponde de Oliveira Pinto and Jared Kaplan and Harri Edwards and Yuri Burda and Nicholas Joseph and Greg Brockman and Alex Ray and Raul Puri and Gretchen Krueger and Michael Petrov and Heidy Khlaaf and Girish Sastry and Pamela Mishkin and Brooke Chan and Scott Gray and Nick Ryder and Mikhail Pavlov and Alethea Power and Lukasz Kaiser and Mohammad Bavarian and Clemens Winter and Philippe Tillet and Felipe Petroski Such and Dave Cummings and Matthias Plappert and Fotios Chantzis and Elizabeth Barnes and Ariel Herbert-Voss and William Hebgen Guss and Alex Nichol and Alex Paino and Nikolas Tezak and Jie Tang and Igor Babuschkin and Suchir Balaji and Shantanu Jain and William Saunders and Christopher Hesse and Andrew N. Carr and Jan Leike and Josh Achiam and Vedant Misra and Evan Morikawa and Alec Radford and Matthew Knight and Miles Brundage and Mira Murati and Katie Mayer and Peter Welinder and Bob McGrew and Dario Amodei and Sam McCandlish and Ilya Sutskever and Wojciech Zaremba},
      year={2021},
      eprint={2107.03374},
      archivePrefix={arXiv},
      primaryClass={cs.LG},
      url={https://arxiv.org/abs/2107.03374}, 
}

@misc{hendrycks2021measuringmathematicalproblemsolving,
      title={Measuring Mathematical Problem Solving With the MATH Dataset}, 
      author={Dan Hendrycks and Collin Burns and Saurav Kadavath and Akul Arora and Steven Basart and Eric Tang and Dawn Song and Jacob Steinhardt},
      year={2021},
      eprint={2103.03874},
      archivePrefix={arXiv},
      primaryClass={cs.LG},
      url={https://arxiv.org/abs/2103.03874}, 
}

@misc{wang2024mmluprorobustchallengingmultitask,
      title={MMLU-Pro: A More Robust and Challenging Multi-Task Language Understanding Benchmark}, 
      author={Yubo Wang and Xueguang Ma and Ge Zhang and Yuansheng Ni and Abhranil Chandra and Shiguang Guo and Weiming Ren and Aaran Arulraj and Xuan He and Ziyan Jiang and Tianle Li and Max Ku and Kai Wang and Alex Zhuang and Rongqi Fan and Xiang Yue and Wenhu Chen},
      year={2024},
      eprint={2406.01574},
      archivePrefix={arXiv},
      primaryClass={cs.CL},
      url={https://arxiv.org/abs/2406.01574}, 
}

@misc{patil2023gorillalargelanguagemodel,
      title={Gorilla: Large Language Model Connected with Massive APIs}, 
      author={Shishir G. Patil and Tianjun Zhang and Xin Wang and Joseph E. Gonzalez},
      year={2023},
      eprint={2305.15334},
      archivePrefix={arXiv},
      primaryClass={cs.CL},
      url={https://arxiv.org/abs/2305.15334}, 
}

@misc{shridhar2021alfworldaligningtextembodied,
      title={ALFWorld: Aligning Text and Embodied Environments for Interactive Learning}, 
      author={Mohit Shridhar and Xingdi Yuan and Marc-Alexandre Côté and Yonatan Bisk and Adam Trischler and Matthew Hausknecht},
      year={2021},
      eprint={2010.03768},
      archivePrefix={arXiv},
      primaryClass={cs.CL},
      url={https://arxiv.org/abs/2010.03768}, 
}

@misc{suzgun2025dynamiccheatsheettesttimelearning,
      title={Dynamic Cheatsheet: Test-Time Learning with Adaptive Memory}, 
      author={Mirac Suzgun and Mert Yuksekgonul and Federico Bianchi and Dan Jurafsky and James Zou},
      year={2025},
      eprint={2504.07952},
      archivePrefix={arXiv},
      primaryClass={cs.LG},
      url={https://arxiv.org/abs/2504.07952}, 
}

@misc{wang2024agentworkflowmemory,
      title={Agent Workflow Memory}, 
      author={Zora Zhiruo Wang and Jiayuan Mao and Daniel Fried and Graham Neubig},
      year={2024},
      eprint={2409.07429},
      archivePrefix={arXiv},
      primaryClass={cs.CL},
      url={https://arxiv.org/abs/2409.07429}, 
}

@misc{zhang2025gmemorytracinghierarchicalmemory,
      title={G-Memory: Tracing Hierarchical Memory for Multi-Agent Systems}, 
      author={Guibin Zhang and Muxin Fu and Guancheng Wan and Miao Yu and Kun Wang and Shuicheng Yan},
      year={2025},
      eprint={2506.07398},
      archivePrefix={arXiv},
      primaryClass={cs.MA},
      url={https://arxiv.org/abs/2506.07398}, 
}

@misc{zhao2024expelllmagentsexperiential,
      title={ExpeL: LLM Agents Are Experiential Learners}, 
      author={Andrew Zhao and Daniel Huang and Quentin Xu and Matthieu Lin and Yong-Jin Liu and Gao Huang},
      year={2024},
      eprint={2308.10144},
      archivePrefix={arXiv},
      primaryClass={cs.LG},
      url={https://arxiv.org/abs/2308.10144}, 
}

@article{wei2025evo,
  title={Evo-memory: Benchmarking llm agent test-time learning with self-evolving memory},
  author={Wei, Tianxin and Sachdeva, Noveen and Coleman, Benjamin and He, Zhankui and Bei, Yuanchen and Ning, Xuying and Ai, Mengting and Li, Yunzhe and He, Jingrui and Chi, Ed H and others},
  journal={arXiv preprint arXiv:2511.20857},
  year={2025}
}

@article{fang2025memp,
  title={Memp: Exploring agent procedural memory},
  author={Fang, Runnan and Liang, Yuan and Wang, Xiaobin and Wu, Jialong and Qiao, Shuofei and Xie, Pengjun and Huang, Fei and Chen, Huajun and Zhang, Ningyu},
  journal={arXiv preprint arXiv:2508.06433},
  year={2025}
}

@article{xiang2026systematic,
  title={A Systematic Survey of Self-Evolving Agents: From Model-Centric to Environment-Driven Co-Evolution},
  author={Xiang, Zhishang and Yang, Chengyi and Chen, Zerui and Wei, Zhimin and Tang, Yunbo and Teng, Zongpei and Peng, Zexi and Li, Zongxia and Huang, Chengsong and He, Yicheng and others},
  year={2026},
  publisher={TechRxiv}
}

@article{zhou2025memento,
  title={Memento: Fine-tuning llm agents without fine-tuning llms},
  author={Zhou, Huichi and Chen, Yihang and Guo, Siyuan and Yan, Xue and Lee, Kin Hei and Wang, Zihan and Lee, Ka Yiu and Zhang, Guchun and Shao, Kun and Yang, Linyi and others},
  journal={arXiv preprint arXiv:2508.16153},
  year={2025}
}

@article{ho2025arcmemo,
  title={Arcmemo: Abstract reasoning composition with lifelong llm memory},
  author={Ho, Matthew and Si, Chen and Feng, Zhaoxiang and Yu, Fangxu and Yang, Yichi and Liu, Zhijian and Hu, Zhiting and Qin, Lianhui},
  journal={arXiv preprint arXiv:2509.04439},
  year={2025}
}

@article{wang2025reinforcement,
  title={Reinforcement learning for self-improving agent with skill library},
  author={Wang, Jiongxiao and Yan, Qiaojing and Wang, Yawei and Tian, Yijun and Mishra, Soumya Smruti and Xu, Zhichao and Gandhi, Megha and Xu, Panpan and Cheong, Lin Lee},
  journal={arXiv preprint arXiv:2512.17102},
  year={2025}
}

@article{zheng2025skillweaver,
  title={Skillweaver: Web agents can self-improve by discovering and honing skills},
  author={Zheng, Boyuan and Fatemi, Michael Y and Jin, Xiaolong and Wang, Zora Zhiruo and Gandhi, Apurva and Song, Yueqi and Gu, Yu and Srinivasa, Jayanth and Liu, Gaowen and Neubig, Graham and others},
  journal={arXiv preprint arXiv:2504.07079},
  year={2025}
}

@article{agrawal2025gepa,
  title={Gepa: Reflective prompt evolution can outperform reinforcement learning},
  author={Agrawal, Lakshya A and Tan, Shangyin and Soylu, Dilara and Ziems, Noah and Khare, Rishi and Opsahl-Ong, Krista and Singhvi, Arnav and Shandilya, Herumb and Ryan, Michael J and Jiang, Meng and others},
  journal={arXiv preprint arXiv:2507.19457},
  year={2025}
}

@article{lopez2017gradient,
  title={Gradient episodic memory for continual learning},
  author={Lopez-Paz, David and Ranzato, Marc'Aurelio},
  journal={Advances in neural information processing systems},
  volume={30},
  year={2017}
}

@inproceedings{wu2022pretrained,
  title={Pretrained Language Model in Continual Learning: A Comparative Study},
  author={Wu, Tongtong and Caccia, Massimo and Li, Zhuang and Li, Yuan Fang and Qi, Guilin and Haffari, Gholamreza},
  booktitle={International Conference on Learning Representations 2022},
  year={2022},
  organization={OpenReview}
}

@misc{qwen3technicalreport,
      title={Qwen3 Technical Report}, 
      author={Qwen Team},
      year={2025},
      eprint={2505.09388},
      archivePrefix={arXiv},
      primaryClass={cs.CL},
      url={https://arxiv.org/abs/2505.09388}, 
}

@misc{minimax2026m27,
  title        = {{MiniMax M2.7}: Early Echoes of Self-Evolution},
  author       = {{MiniMax}},
  year         = {2026},
  howpublished = {\url{https://www.minimax.io/news/minimax-m27-en}},
  note         = {Accessed: 2026-05-04}
}

@article{xia2026skillrl,
  title={Skillrl: Evolving agents via recursive skill-augmented reinforcement learning},
  author={Xia, Peng and Chen, Jianwen and Wang, Hanyang and Liu, Jiaqi and Zeng, Kaide and Wang, Yu and Han, Siwei and Zhou, Yiyang and Zhao, Xujiang and Chen, Haifeng and others},
  journal={arXiv preprint arXiv:2602.08234},
  year={2026}
}

@inproceedings{reimers2019sentence,
  title={Sentence-bert: Sentence embeddings using siamese bert-networks},
  author={Reimers, Nils and Gurevych, Iryna},
  booktitle={Proceedings of the 2019 conference on empirical methods in natural language processing and the 9th international joint conference on natural language processing (EMNLP-IJCNLP)},
  pages={3982--3992},
  year={2019}
}

@article{fang2025comprehensive,
  title={A comprehensive survey of self-evolving ai agents: A new paradigm bridging foundation models and lifelong agentic systems},
  author={Fang, Jinyuan and Peng, Yanwen and Zhang, Xi and Wang, Yingxu and Yi, Xinhao and Zhang, Guibin and Xu, Yi and Wu, Bin and Liu, Siwei and Li, Zihao and others},
  journal={arXiv preprint arXiv:2508.07407},
  year={2025}
  }

@article{madaan2023self,
  title={Self-refine: Iterative refinement with self-feedback},
  author={Madaan, Aman and Tandon, Niket and Gupta, Prakhar and Hallinan, Skyler and Gao, Luyu and Wiegreffe, Sarah and Alon, Uri and Dziri, Nouha and Prabhumoye, Shrimai and Yang, Yiming and others},
  journal={Advances in neural information processing systems},
  volume={36},
  pages={46534--46594},
  year={2023}
}

@article{chhikara2025mem0,
  title={Mem0: Building production-ready ai agents with scalable long-term memory},
  author={Chhikara, Prateek and Khant, Dev and Aryan, Saket and Singh, Taranjeet and Yadav, Deshraj},
  journal={arXiv preprint arXiv:2504.19413},
  year={2025}
}

@inproceedings{zhong2024memorybank,
  title={Memorybank: Enhancing large language models with long-term memory},
  author={Zhong, Wanjun and Guo, Lianghong and Gao, Qiqi and Ye, He and Wang, Yanlin},
  booktitle={Proceedings of the AAAI conference on artificial intelligence},
  volume={38},
  number={17},
  pages={19724--19731},
  year={2024}
}

@article{xu2025mem,
  title={A-mem: Agentic memory for llm agents},
  author={Xu, Wujiang and Liang, Zujie and Mei, Kai and Gao, Hang and Tan, Juntao and Zhang, Yongfeng},
  journal={arXiv preprint arXiv:2502.12110},
  year={2025}
}

@article{chen2025swe,
  title={Swe-exp: Experience-driven software issue resolution},
  author={Chen, Silin and Lin, Shaoxin and Shi, Yuling and Lian, Heng and Gu, Xiaodong and Yun, Longfei and Chen, Dong and Cao, Lin and Liu, Jiyang and Xia, Nu and others},
  journal={arXiv preprint arXiv:2507.23361},
  year={2025}
}

@inproceedings{wang2025far,
  title={How Far Can LLMs Improve from Experience? Measuring Test-Time Learning Ability in LLMs with Human Comparison},
  author={Wang, Jiayin and Guo, Zhiqiang and Ma, Weizhi and Zhang, Min},
  booktitle={Proceedings of the 2025 Conference on Empirical Methods in Natural Language Processing},
  pages={25688--25702},
  year={2025}
}

@article{gao2025survey,
  title={A Survey of Self-Evolving Agents: What, When, How, and Where to Evolve on the Path to Artificial Super Intelligence},
  author={Gao, Huan-ang and Geng, Jiayi and Hua, Wenyue and Hu, Mengkang and Juan, Xinzhe and Liu, Hongzhang and Liu, Shilong and Qiu, Jiahao and Qi, Xuan and Wu, Yiran and others},
  journal={arXiv preprint arXiv:2507.21046},
  year={2025}
}

@article{feng2025get,
  title={Get experience from practice: Llm agents with record \& replay},
  author={Feng, Erhu and Zhou, Wenbo and Liu, Zibin and Chen, Le and Dong, Yunpeng and Zhang, Cheng and Zhao, Yisheng and Du, Dong and Hua, Zhichao and Xia, Yubin and others},
  journal={arXiv preprint arXiv:2505.17716},
  year={2025}
}

@article{tang2025agent,
  title={Agent kb: Leveraging cross-domain experience for agentic problem solving},
  author={Tang, Xiangru and Qin, Tianrui and Peng, Tianhao and Zhou, Ziyang and Shao, Daniel and Du, Tingting and Wei, Xinming and Xia, Peng and Wu, Fang and Zhu, He and others},
  journal={arXiv preprint arXiv:2507.06229},
  year={2025}
}

@article{ouyang2025reasoningbank,
  title={Reasoningbank: Scaling agent self-evolving with reasoning memory},
  author={Ouyang, Siru and Yan, Jun and Hsu, I and Chen, Yanfei and Jiang, Ke and Wang, Zifeng and Han, Rujun and Le, Long T and Daruki, Samira and Tang, Xiangru and others},
  journal={arXiv preprint arXiv:2509.25140},
  year={2025}
}

@article{wu2025evolver,
  title={Evolver: Self-evolving llm agents through an experience-driven lifecycle},
  author={Wu, Rong and Wang, Xiaoman and Mei, Jianbiao and Cai, Pinlong and Fu, Daocheng and Yang, Cheng and Wen, Licheng and Yang, Xuemeng and Shen, Yufan and Wang, Yuxin and others},
  journal={arXiv preprint arXiv:2510.16079},
  year={2025}
}

@inproceedings{biesialska2020continual,
  title={Continual lifelong learning in natural language processing: A survey},
  author={Biesialska, Magdalena and Biesialska, Katarzyna and Costa-Jussa, Marta R},
  booktitle={Proceedings of the 28th international conference on computational linguistics},
  pages={6523--6541},
  year={2020}
}

@article{kirkpatrick2017overcoming,
  title={Overcoming catastrophic forgetting in neural networks},
  author={Kirkpatrick, James and Pascanu, Razvan and Rabinowitz, Neil and Veness, Joel and Desjardins, Guillaume and Rusu, Andrei A and Milan, Kieran and Quan, John and Ramalho, Tiago and Grabska-Barwinska, Agnieszka and others},
  journal={Proceedings of the national academy of sciences},
  volume={114},
  number={13},
  pages={3521--3526},
  year={2017},
  publisher={National Academy of Sciences}
}

@article{wang2024comprehensive,
  title={A comprehensive survey of continual learning: Theory, method and application},
  author={Wang, Liyuan and Zhang, Xingxing and Su, Hang and Zhu, Jun},
  journal={IEEE transactions on pattern analysis and machine intelligence},
  volume={46},
  number={8},
  pages={5362--5383},
  year={2024},
  publisher={IEEE}
}

@article{qi2023fine,
  title={Fine-tuning aligned language models compromises safety, even when users do not intend to!},
  author={Qi, Xiangyu and Zeng, Yi and Xie, Tinghao and Chen, Pin-Yu and Jia, Ruoxi and Mittal, Prateek and Henderson, Peter},
  journal={arXiv preprint arXiv:2310.03693},
  year={2023}
}

@article{wang2023trace,
  title={Trace: A comprehensive benchmark for continual learning in large language models},
  author={Wang, Xiao and Zhang, Yuansen and Chen, Tianze and Gao, Songyang and Jin, Senjie and Yang, Xianjun and Xi, Zhiheng and Zheng, Rui and Zou, Yicheng and Gui, Tao and others},
  journal={arXiv preprint arXiv:2310.06762},
  year={2023}
}

@article{chaudhry2019tiny,
  title={On tiny episodic memories in continual learning},
  author={Chaudhry, Arslan and Rohrbach, Marcus and Elhoseiny, Mohamed and Ajanthan, Thalaiyasingam and Dokania, Puneet K and Torr, Philip HS and Ranzato, Marc'Aurelio},
  journal={arXiv preprint arXiv:1902.10486},
  year={2019}
}

@article{chaudhry2018efficient,
  title={Efficient lifelong learning with a-gem},
  author={Chaudhry, Arslan and Ranzato, Marc'Aurelio and Rohrbach, Marcus and Elhoseiny, Mohamed},
  journal={arXiv preprint arXiv:1812.00420},
  year={2018}
}

@article{wu2024continual,
  title={Continual learning for large language models: A survey},
  author={Wu, Tongtong and Luo, Linhao and Li, Yuan-Fang and Pan, Shirui and Vu, Thuy-Trang and Haffari, Gholamreza},
  journal={arXiv preprint arXiv:2402.01364},
  year={2024}
}

@article{shi2025continual,
  title={Continual learning of large language models: A comprehensive survey},
  author={Shi, Haizhou and Xu, Zihao and Wang, Hengyi and Qin, Weiyi and Wang, Wenyuan and Wang, Yibin and Wang, Zifeng and Ebrahimi, Sayna and Wang, Hao},
  journal={ACM Computing Surveys},
  volume={58},
  number={5},
  pages={1--42},
  year={2025},
  publisher={ACM New York, NY}
}

@article{li2025memos,
  title={Memos: An operating system for memory-augmented generation (mag) in large language models},
  author={Li, Zhiyu and Song, Shichao and Wang, Hanyu and Niu, Simin and Chen, Ding and Yang, Jiawei and Xi, Chenyang and Lai, Huayi and Zhao, Jihao and Wang, Yezhaohui and others},
  journal={arXiv preprint arXiv:2505.22101},
  year={2025}
}

@article{liang2025sage,
  title={Sage: Self-evolving agents with reflective and memory-augmented abilities},
  author={Liang, Xuechen and Tao, Meiling and Xia, Yinghui and Wang, Jianhui and Li, Kun and Wang, Yijin and He, Yangfan and Yang, Jingsong and Shi, Tianyu and Wang, Yuantao and others},
  journal={Neurocomputing},
  volume={647},
  pages={130470},
  year={2025},
  publisher={Elsevier}
}

@article{zheng2023synapse,
  title={Synapse: Trajectory-as-exemplar prompting with memory for computer control},
  author={Zheng, Longtao and Wang, Rundong and Wang, Xinrun and An, Bo},
  journal={arXiv preprint arXiv:2306.07863},
  year={2023}
}
